\documentclass{article}
\usepackage{iclr2027_conference,times}
\usepackage{graphicx}
\usepackage{xcolor}
\usepackage{booktabs,tabularx,array,longtable,multirow}
\usepackage{amsmath,amsfonts,amssymb,mathtools}
\usepackage{placeins,float}
\usepackage{needspace}
\usepackage{wrapfig,capt-of}
\usepackage{algorithm,algorithmic}
\usepackage{colortbl}
\usepackage[most]{tcolorbox}
\usepackage[hyphens]{url}
\usepackage{hyperref}
\hypersetup{hidelinks,pdfauthor={Rui Zou, Mengqi Wei, Xin Zhao, Ji-Rong Wen},pdftitle={Agent-Guided Process Verification for Mathematical Self-Correction}}
\newtcolorbox{agfvexamplebox}{
  enhanced,
  breakable,
  width=\linewidth,
  colback=white,
  colframe=black,
  boxrule=0.45pt,
  arc=1pt,
  outer arc=1pt,
  left=6pt,
  right=6pt,
  top=6pt,
  bottom=6pt,
  before skip=6pt,
  after skip=6pt
}
\newtcolorbox{breakablecasebox}{
  enhanced,
  breakable,
  width=\linewidth,
  center,
  colback=white,
  colframe=black,
  boxrule=0.4pt,
  arc=0pt,
  outer arc=0pt,
  left=6pt,
  right=6pt,
  top=6pt,
  bottom=6pt,
  before skip=6pt,
  after skip=6pt
}

\newcommand{\scoreup}[1]{\textcolor[HTML]{287A54}{#1}}
\newcommand{\scoredown}[1]{\textcolor[HTML]{B64242}{#1}}

\newcommand{\accgain}[2]{\ensuremath{\text{#1}_{\,\text{\scriptsize #2}}}}
\newcommand{\citepboxed}[1]{\citep{#1}}

\DeclareRobustCommand{\figref}[1]{Fig.~\ref{#1}}

\DeclareRobustCommand{\tabref}[1]{Table~\ref{#1}}

\title{Agent-Guided Process Verification\\for Mathematical Self-Correction}
\author{
Rui Zou$^{1}$ \quad Mengqi Wei$^{2}$ \quad Xin Zhao$^{1}$ \quad Ji-Rong Wen$^{1}$\thanks{Corresponding author.}\\[4pt]
\normalfont $^{1}$Gaoling School of Artificial Intelligence, Renmin University of China\\
\normalfont $^{2}$National Engineering Research Center of Educational Big Data,\\
\normalfont Central China Normal University\\[2pt]
\normalfont \texttt{zouruixyz@ruc.edu.cn, weimengqi@mails.ccnu.edu.cn}\\
\normalfont \texttt{batmanfly@gmail.com, jrwen@ruc.edu.cn}
}
\date{}
\usepackage{etoolbox}
\iclrfinalcopy
\makeatletter
\patchcmd{\@maketitle}
  {\lhead{Published as a conference paper at ICLR 2027}}
  {\lhead{}}
  {}
  {\PackageError{AGPV}{Could not remove the conference header}{Check the preprint layout.}}
\makeatother
\renewcommand{\headrulewidth}{0pt}
\makeatletter
\g@addto@macro\@maketitle{%
  \g@addto@macro\@thanks{%
    \footnotetext[0]{\textbf{Code}\enspace\url{https://github.com/TicusFFF/AGPV}}}}
\makeatother
\begin{document}
\raggedbottom
\maketitle
\begin{abstract}
Large language models have made substantial progress in mathematical reasoning, yet in complex multi-step reasoning, an error at any step may propagate along an incorrect path, and the resulting solution may omit necessary branches. Existing natural-language reflection methods often struggle to locate critical errors precisely, while outcome verification typically checks only whether the available candidate answers satisfy the problem conditions and can therefore overlook necessary branches. To address these limitations, we propose Agent-Guided Process Verification (AGPV) for mathematical self-correction. AGPV decomposes the current solution into an ordered sequence of key claims, verifies them individually while invoking a shared mathematical toolbox when needed, and stops upon detecting the first error. A summarization agent converts the failure evidence into a diagnosis that guides a correction agent to re-derive the subsequent solution from that point; AGPV then adapts the next verification iteration to the revised solution. AGPV also verifies solution completeness. We further prove that AGPV guarantees solution correctness and completeness when key claims provide sufficient coverage and local verification is reliable, provided that all key claims pass verification. The implementation compresses long working contexts and reuses successful tool results within each problem. We evaluate AGPV on five challenging mathematical reasoning datasets with four models from DeepSeek, Qwen, and GPT. AGPV achieves an average accuracy of 63.8\%, outperforming the strongest among the representative baselines evaluated (50.3\%) by 13.5 percentage points, with higher repair coverage than the programmatic-verification baselines.
\end{abstract}

\newlength{\agpvsavedtextfloatsep}
\newlength{\agpvsavedfloatsep}
\newlength{\agpvsavedintextsep}
\newlength{\agpvsavedabovecaptionskip}
\setlength{\agpvsavedtextfloatsep}{\textfloatsep}
\setlength{\agpvsavedfloatsep}{\floatsep}
\setlength{\agpvsavedintextsep}{\intextsep}
\setlength{\agpvsavedabovecaptionskip}{\abovecaptionskip}
\setlength{\textfloatsep}{6pt plus 2pt minus 2pt}
\setlength{\floatsep}{6pt plus 2pt minus 2pt}
\setlength{\intextsep}{5pt plus 2pt minus 2pt}
\setlength{\abovecaptionskip}{4pt}

\section{Introduction}

Large language models have made substantial progress in mathematical reasoning~\citepboxed{wei2022chain,lewkowycz2022minerva,guo2025deepseekr1}. However, in complex multi-step mathematical problems, an error at one step may cause the subsequent reasoning to proceed along an incorrect path and ultimately produce a wrong answer~\citepboxed{uesato2022process,lightman2023verify}. Effective error checking and self-correction are therefore essential to improving the reliability of mathematical reasoning in large language models.
Existing approaches primarily include natural-language reflection, iterative refinement, structured checking, and program- or tool-augmented verification~\citepboxed{pan-etal-2024-automatically,gou2024critic}. For example, Self-Refine~\citepboxed{madaan2023selfrefine} and Reflexion~\citepboxed{shinn2023reflexion} use natural-language feedback to support reflection and iterative refinement, while CheckList~\citepboxed{cook2024ticking} organizes evaluation and correction through generated checklists. ProgCo~\citepboxed{song-etal-2025-progco}, by contrast, employs a programmatic verification mechanism to perform outcome verification from candidate answers.

\begin{wrapfigure}{r}{0.485\textwidth}
\centering
\includegraphics[width=\linewidth]{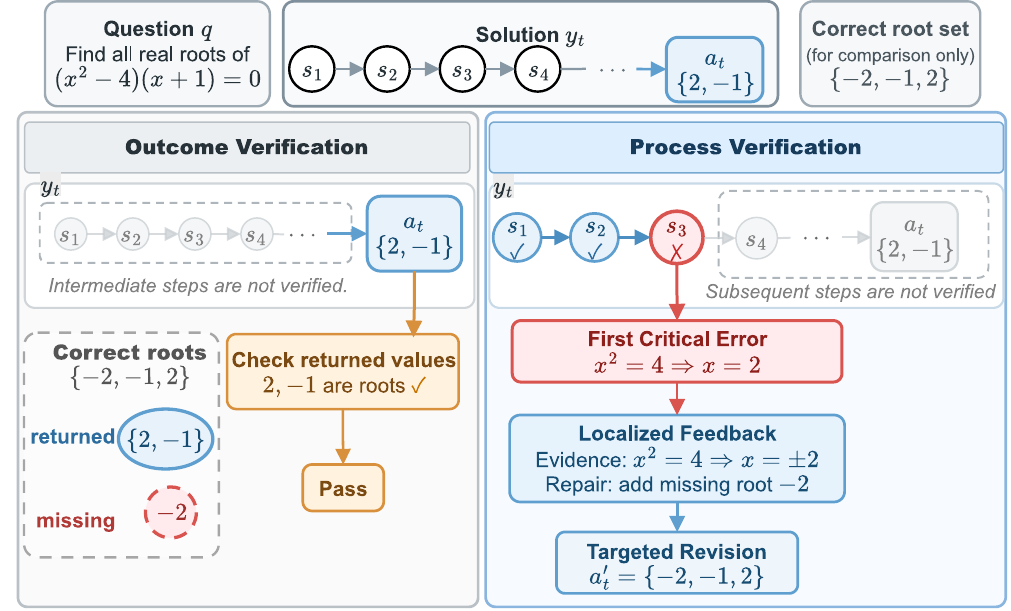}
\caption{Comparison between outcome verification and AGPV process verification on an incomplete-root-set example. Here, \(s_1,s_2,\ldots\) denote the ordered solution steps in the current solution \(y_t\).}
\label{fig:agpv_introduction_comparison}
\end{wrapfigure}

These approaches nevertheless have two limitations. First, holistic natural-language reflection has difficulty locating errors precisely. Even when it identifies several suspicious points, it may fail to find the first critical error that actually invalidates the subsequent reasoning, leading to unfocused, wholesale rewriting~\citepboxed{huang2024selfcorrect,tyen-etal-2024-llms}. Second, outcome verification starting from candidate answers typically focuses on whether the available results satisfy the problem conditions~\citepboxed{uesato2022process,lightman2023verify}. However, it does not explicitly represent whether all valid solutions and necessary branches have been covered. It may therefore miss incompleteness errors such as incomplete solution sets or omitted branches. As shown in Fig.~\ref{fig:agpv_introduction_comparison}, the problem asks for all real roots of an equation. Substituting the available results \(\{2,-1\}\) into the original equation verifies both roots, yet the solution set remains incomplete. In contrast, checking the key claim ``whether all solutions to \(x^2=4\) have been found'' along the solution process reveals the omission and precisely localizes the error.

The first critical error in mathematical reasoning resembles the first falling domino: subsequent errors often emerge through a cascading effect. It is therefore important to localize the first critical error in the solution process promptly and begin correction from that point. Tyen et al. show that providing an error location can facilitate correction~\citepboxed{tyen-etal-2024-llms}. Motivated by this, AGPV constructs and sequentially verifies key claims along the solution process to localize the first claim that compromises the correctness or completeness of subsequent reasoning. Step-level process verification provides fine-grained signals over intermediate reasoning~\citepboxed{lightman2023verify,wang-etal-2024-math}. AGPV uses this granularity to check whether necessary branches are complete as the reasoning unfolds, thereby verifying completeness within the same process.

Accordingly, we propose AGPV. Here, process verification means converting the current solution into an ordered sequence of locally verifiable key claims and checking them in their original reasoning order. We call the method agent-guided because specialized agents dynamically construct and verify solution-specific key claims, summarize the first failed claim together with supporting evidence, and use this diagnosis to guide localized correction of the subsequent reasoning. Within the AGPV framework, a generation agent first parses the current solution and constructs an ordered sequence of key claims. A verification agent then checks these claims sequentially and, when necessary, invokes a shared mathematical toolbox to obtain mathematical evidence such as equation-solving and constraint-checking results. Upon encountering the first key claim that fails verification, the framework stops searching and initiates correction. A summarization agent then generates feedback from the failure evidence, guiding a correction agent to re-derive the subsequent solution from that point. Based on whether the answers before and after correction are equivalent, the framework regenerates or adapts the key claims for the revised solution and repeats this process until all key claims pass verification or the iteration limit is reached. We further prove that, under ideal conditions in which the key claims provide sufficient coverage and local verification is reliable, AGPV guarantees the correctness and completeness of the final solution, provided that all key claims pass verification.

We evaluate AGPV on five challenging mathematical reasoning datasets with four models from DeepSeek, Qwen, and GPT. AGPV achieves an average accuracy of \(63.8\%\), outperforming the strongest evaluated baseline (\(50.3\%\)) by \(13.5\) percentage points. Correction analysis shows that it repairs more initially incorrect answers, with particularly pronounced gains on problems that the initial models solve less reliably.

Our contributions are as follows: (1) We propose AGPV, which uses process verification to localize the first key claim that fails verification and guides the model to re-derive the subsequent solution from that point; (2) AGPV naturally incorporates solution-completeness checking into process verification without requiring a separate completeness-checking stage, enabling it to identify incompleteness errors. We further provide a corresponding conditional reliability proof; and (3) We develop a multi-agent implementation of AGPV with a shared mathematical toolbox, iterative key-claim adaptation, working-context compression, and within-problem reuse of successful tool results, and validate its effectiveness across multiple mathematical datasets and model settings.

\section{Related Work}
\label{sec:related_work}

This work relates to three research areas: self-correction in large language models, process verification, and tool-augmented mathematical reasoning.

\noindent\textbf{Self-Correction in Large Language Models.}
Prior work improves test-time outputs through natural-language feedback, reflection, structured checking, and multi-path sampling. Self-Refine~\citepboxed{madaan2023selfrefine} iteratively generates feedback and revises its responses; Reflexion~\citepboxed{shinn2023reflexion} uses linguistic feedback to summarize prior failures; TICK~\citepboxed{cook2024ticking} organizes evaluation and correction with automatically generated checklists; and Self-Consistency~\citepboxed{wang2022selfconsistency} selects a consistent answer across multiple reasoning paths.
Recent studies further improve self-verification and self-correction through training. S\(^2\)R~\citepboxed{ma2025s2r} uses reinforcement learning to train models for self-verification and correction, while SPOC~\citepboxed{zhao2025spoc} interleaves solution generation and verification in a single inference pass to trigger spontaneous correction.
Collectively, these studies show that feedback and test-time iteration can improve large language models' reasoning outcomes.

\noindent\textbf{Process Verification.}
Verifiers commonly evaluate, rank, or filter candidate solutions. Early mathematical reasoning work used outcome verifiers~\citepboxed{cobbe2021gsm8k} to select more reliable outputs from multiple candidates.
Uesato et al.~\citepboxed{uesato2022process} compared outcome-based and process-based feedback for mathematical reasoning. Lightman et al.~\citepboxed{lightman2023verify} introduced PRM800K with step-level annotations and used process supervision to train process reward models. Math-Shepherd~\citepboxed{wang-etal-2024-math} automated process-supervision data construction, assigned rewards to intermediate reasoning steps, and applied them to candidate ranking and reinforcement learning.
For test-time self-correction, ProgCo~\citepboxed{song-etal-2025-progco} applies program-guided verification to candidate solutions and generates corrective feedback from execution results.
Together, these studies extend verification from final outcomes to intermediate reasoning and provide a foundation for identifying process errors.

\begin{figure}[t]
\centering
\includegraphics[width=0.97\textwidth]{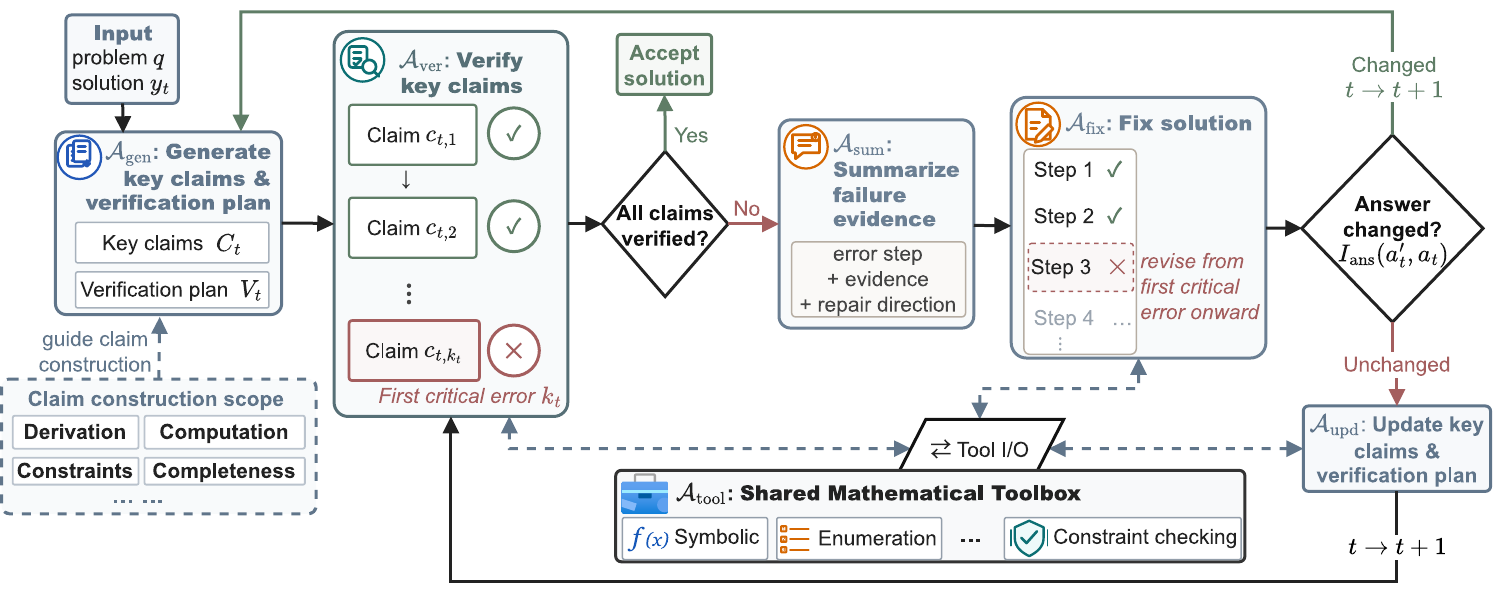}
\caption{The AGPV workflow: sequential key-claim verification, first-error localization, failure-evidence summarization, solution correction, and verification-plan regeneration or updating.}
\label{fig:agfv_framework}
\end{figure}

\noindent\textbf{Tool-Augmented Mathematical Reasoning.}
Tool-augmented reasoning combines programs, code interpreters, and specialized mathematical tools to improve the reliability of language models' symbolic derivations and exact computation. PAL~\citepboxed{gao2023pal} translates problems into programs executed by a Python interpreter; Program-of-Thoughts~\citepboxed{chen2023pot} separates computation from natural-language reasoning; and ToRA~\citepboxed{gou2024tora} alternates between natural-language reasoning and external tool calls to solve complex mathematical problems.
ReAct~\citepboxed{yao2023react} coordinates reasoning with external actions, while Toolformer~\citepboxed{schick2023toolformer} enables language models to select and invoke tools autonomously.
Recent work such as AgentMath~\citepboxed{luo2025agentmath} and R1-Code-Interpreter~\citepboxed{chen2025r1codeinterpreter} further integrates code interpreters and mathematical tools with agent training for complex mathematical reasoning.

Overall, prior work improves mathematical reasoning through test-time self-correction, process supervision, and tool augmentation. AGPV builds on these directions by constructing ordered key claims from the current solution and problem objective, locating the first failed claim, checking completeness claim by claim, and using tool evidence to focus correction on the subsequent solution.

\section{Method}

Figure~\ref{fig:agfv_framework} presents the overall AGPV framework. Given a problem \(q\), the current solution \(y_t\) at iteration \(t\), and its extracted answer \(a_t\), AGPV generates an ordered sequence of locally verifiable key claims and checks them in an order aligned with the original reasoning process. If all claims pass, the system accepts the current answer. Otherwise, it identifies the first failed claim as the first critical error, summarizes the failure evidence, and corrects the subsequent solution from that point.
After correction, the answer-equivalence controller compares the answers before and after correction to decide whether the next iteration should regenerate or update the key-claim sequence and verification plan. Key-claim verification, solution correction, and updates to the key claims and verification plan may invoke the shared mathematical toolbox for exact computation. The following subsections describe key-claim generation and verification, solution correction, and verification-plan updates. Algorithm~\ref{alg:agfv} in Appendix~\ref{app:overall_algorithm} provides the complete AGPV control procedure.

\par\begingroup\noindent\ignorespaces
Table~\ref{tab:agpv_added_prompts} in Appendix~\ref{app:overall_algorithm} lists the prompts for the six agents in Figure~\ref{fig:agfv_framework}.
\par\endgroup

\subsection{Generating Key Claims}

AGPV verifies a sequence of key claims jointly determined by the current solution and the problem objective. Given problem \(q\), current solution \(y_t\), and extracted answer \(a_t\) at iteration \(t\), \(\mathcal{A}_{\mathrm{gen}}\) generates the key-claim sequence and its verification plan:
\begin{equation}
\label{eq:agfv-generate-claims}
(C_t,V_t)=\mathcal{A}_{\mathrm{gen}}(q,y_t,a_t),
\qquad C_t=(c_{t,1},c_{t,2},\ldots,c_{t,n_{\scriptstyle t}}).
\end{equation}
Here, the ordered key-claim sequence \(C_t\) specifies what to verify, and \(V_t\) specifies how to verify it along the current solution. The term \(c_{t,i}\) is the \(i\)-th verifiable mathematical claim at iteration \(t\), and \(n_t\) is the number of key claims in that iteration. These claims may come from explicit reasoning steps or be induced by the problem objective, including algebraic simplification, equation transformation, numerical computation, constraint derivation, and completeness requirements.

Key claims are distinct from solution steps. Solution steps are the original reasoning units that appear in their written order in the current solution \(y_t\), whereas a key claim \(c_{t,i}\) is an independently verifiable mathematical proposition extracted or constructed from one or more solution steps and the problem objective. Their mapping need not be one-to-one: one step may yield multiple claims, several steps may support one claim, and the problem objective may induce completeness claims that are not stated explicitly in the current solution. AGPV orders key claims by the original solution process, while \(V_t\) specifies their verification.
The verification plan \(V_t\) specifies the order in which the key claims are verified, their local verification objectives, the tool types that may be required, and the evidence to retain during verification. AGPV decomposes the current solution into key claims ordered by the original reasoning process, enabling subsequent verification to localize errors to specific reasoning stages rather than issuing a single judgment on the final answer.

AGPV naturally incorporates solution-completeness checks into process verification. For problems requiring all solutions, complete counting, exhaustive enumeration, a complete set, or a global optimum, it converts each requirement into a key claim in \(C_t\). These requirements thus enter the same process-verification chain as local computations, constraint derivations, and intermediate conclusions.

\subsection{Verifying Key Claims}

AGPV verifies the current solution claim by claim and invokes the shared mathematical toolbox agent when exact computation is required, producing traceable tool evidence. Given the key-claim sequence \(C_t\) and verification plan \(V_t\) from the preceding subsection, \(\mathcal{A}_{\mathrm{ver}}\) returns
\begin{equation}
\label{eq:agfv-verification-agent}
(r_t,R_t,k_t)
=
\mathcal{A}_{\mathrm{ver}}(q,y_t,a_t,C_t,V_t).
\end{equation}
Here, \(r_t\in\{0,1\}\) indicates whether verification succeeds at iteration \(t\); \(R_t\) stores the examined key claims, corresponding tool evidence, and pass/fail decisions; and \(k_t\) is the position of the first failed claim, with \(k_t=\varnothing\) when all claims pass. Under the ideal conditions that the key claims sufficiently cover the current solution process, preserve the original reasoning order, and are verified reliably, the first failed claim corresponds to the first critical error. We use this term below.
Specifically, \(\mathcal{A}_{\mathrm{ver}}\) checks the claims in \(C_t\) in the order specified by \(V_t\). For each claim, \(\mathcal{A}_{\mathrm{ver}}\) determines whether the problem conditions and previously verified steps support it. \(\mathcal{A}_{\mathrm{ver}}\) directly returns pass or fail for simple logical or expression-consistency checks. For algebraic simplification, equation solving, enumeration, counting, constraint checking, or exact numerical computation, \(\mathcal{A}_{\mathrm{ver}}\) sends a structured request to the shared mathematical toolbox agent and uses the result as verification evidence.

\paragraph{Shared Mathematical Toolbox Agent.}
Let \(\mathbb T\) denote the available tools for symbolic simplification, equation solving, enumeration and counting, constraint checking, and exact computation. When reliable computation is needed during key-claim verification, solution correction, or updates to the key claims and verification plan, the corresponding agent generates a structured request \(\tau_t^{s}\), where
\(s\in\{\mathrm{ver},\mathrm{fix},\mathrm{upd}\}\).
\(\mathcal{A}_{\mathrm{tool}}\) selects a tool and generates an executable call:
\begin{equation}
\label{eq:agpv-tool-call}
\kappa_t^{s}=\mathcal A_{\mathrm{tool}}(\tau_t^{s};\mathbb T),
\qquad z_t^{s}=\operatorname{Run}(\kappa_t^{s}).
\end{equation}
Here, \(\kappa_t^{s}\) is the generated tool call, \(\operatorname{Run}\) executes it on the tool backend, and \(z_t^{s}\) is the returned result. Equation~\eqref{eq:agpv-tool-call} may be repeated as needed within a stage. Verification results enter \(R_t\); correction results support new computations or derivations; and update results help inspect the updated verification targets. \(\mathcal A_{\mathrm{sum}}\) does not call the toolbox directly but summarizes the tool evidence already stored in \(R_t\).
Our implementation supports SymPy~\citepboxed{meurer2017sympy}, Z3~\citepboxed{demoura2008z3}, and Python tools for exact numerical computation and enumeration.

If all key claims pass verification, \(r_t=1\) and \(k_t=\varnothing\), and the system accepts the current solution and extracted answer. Otherwise, AGPV identifies the first failed claim as the first critical error:
\begin{equation}
\label{eq:agfv-first-failed-claim}
k_t=\min\{i\mid c_{t,i}\ \text{fails verification}\},
\qquad r_t=0.
\end{equation}
Verification stops immediately, and \(R_t\) retains the verification trace from the first claim through the first failed claim. Stopping at the first critical error is necessary because later reasoning may already depend on an erroneous premise, and continuing to inspect subsequent claims can obscure the actual source of the error.

\subsection{Correcting the Solution and Updating the Verification Plan}

When \(r_t=0\), AGPV enters feedback-based correction, involving \(\mathcal{A}_{\mathrm{sum}}\), \(\mathcal{A}_{\mathrm{fix}}\), \(\mathcal{A}_{\mathrm{upd}}\), and an answer-equivalence controller. The agents respectively summarize failure evidence into feedback, correct the solution, and update the key-claim sequence and verification plan when the answers are equivalent. Based on answer equivalence, the controller directs \(\mathcal{A}_{\mathrm{gen}}\) to regenerate the key claims and verification plan for the next iteration or \(\mathcal{A}_{\mathrm{upd}}\) to update them.

First, \(\mathcal{A}_{\mathrm{sum}}\) uses the current verification plan \(V_t\), the verification trace \(R_t\), and the position \(k_t\) of the first failed claim to summarize the failure evidence; \(\mathcal{A}_{\mathrm{fix}}\) then corrects \(y_t\) using \(F_t\):
\begin{align}
F_t &= \mathcal{A}_{\mathrm{sum}}(q,y_t,a_t,V_t,R_t,k_t),
\label{eq:agfv-feedback-summary}\\
y'_t &= \mathcal{A}_{\mathrm{fix}}(q,y_t,a_t,F_t).
\label{eq:agfv-answer-fix}
\end{align}
Here, \(F_t\) is natural-language feedback identifying the first critical error, the relevant tool evidence in \(R_t\), why the key claim fails, and where correction should begin. Because the evidence is read directly from \(R_t\), no separate execution-record variable is required.
Here, \(y'_t\) is the corrected candidate solution, and \(a'_t\) is its extracted answer. Guided by the error location in \(F_t\), \(\mathcal{A}_{\mathrm{fix}}\) re-derives the subsequent solution from the first critical error rather than regenerating the entire solution. When recomputation, constraint checking, or verification of a new derivation is required, it may invoke \(\mathcal{A}_{\mathrm{tool}}\) and use the returned result.

After correction, the answer-equivalence controller compares \(a'_t\) and \(a_t\), where \(I_{\mathrm{ans}}(x,y)=1\) denotes equivalence and \(I_{\mathrm{ans}}(x,y)=0\) non-equivalence. If \(I_{\mathrm{ans}}(a'_t,a_t)=0\), the changed answer may make \(C_t\) and \(V_t\) inapplicable. The system sets \(y_{t+1}=y'_t\) and \(a_{t+1}=a'_t\), clears \(C_t,V_t\), and invokes \(\mathcal{A}_{\mathrm{gen}}\) in the next iteration to regenerate \(C_{t+1},V_{t+1}\) from \(q,y_{t+1},a_{t+1}\).

If \(I_{\mathrm{ans}}(a'_t,a_t)=1\), the answers are equivalent, but the revised solution may no longer align with \(C_t,V_t\). Direct reuse risks inconsistency, whereas full regeneration may discard the previous failure diagnosis. Therefore, \(\mathcal{A}_{\mathrm{upd}}\) updates the key-claim sequence and verification plan using \(y'_t,a'_t,C_t,V_t,R_t,F_t\). If \(V_t\) contains missing or invalid tool tasks or conflicts with the corrected solution, \(\mathcal{A}_{\mathrm{upd}}\) may invoke \(\mathcal{A}_{\mathrm{tool}}\) before producing:
\begin{equation}
\label{eq:agfv-update-plan}
\begin{aligned}
(C_{t+1},V_{t+1})
&=
\mathcal{A}_{\mathrm{upd}}(q,y'_t,a'_t,C_t,V_t,R_t,F_t).
\end{aligned}
\end{equation}
The implementation compares extracted answer strings; differently formatted but mathematically equivalent answers may therefore trigger regeneration rather than adaptation.
In both branches, \(y_{t+1}=y'_t\) and \(a_{t+1}=a'_t\), so, if another iteration remains, it verifies the corrected solution. When the iteration limit is reached, AGPV returns the latest corrected candidate without further verification. In the equivalent-answer branch, \(\mathcal{A}_{\mathrm{upd}}\) uses the evidence in \(R_t\) and the diagnosis in \(F_t\) to update \(C_{t+1}\) and \(V_{t+1}\), keeping them consistent with \(y'_t\).

\paragraph{Working-Context Compression and Tool-Result Reuse.}
The implementation maintains a working context for continued verification and correction. When it reaches the configured token threshold, a compression agent summarizes older working content while retaining the fixed instructions and recent interaction context. The compression prompt requests preservation of key computations, tool evidence, failed approaches, and unfinished checks; the full execution records remain available separately. Before feedback summarization and verification-plan updates, long execution traces can likewise be replaced by compact working summaries. For repeated mathematical-tool requests within the same problem, the executor reuses an earlier successful, nonempty result for the same request text (after trimming surrounding whitespace), avoiding repeated code generation and Python execution. Failed or empty results are not cached, and the cache is not shared across problems. These mechanisms support the verification workflow; their individual effects are not isolated by the experiments reported here.

\paragraph{Conditional Reliability.}
The result assumes sufficient coverage of the correctness and completeness requirements and reliable verification of each claim against the problem conditions and previously accepted claims. Intuitively, the argument follows mathematical induction over the ordered claims. The first claim follows from the problem conditions. If the first \(i-1\) claims have been established, reliable local verification shows that the \(i\)-th claim also follows from the problem conditions and those preceding claims. Repeating this step establishes all claims, and sufficient coverage then implies final-answer correctness and completeness.
Let \(P_q\) denote the problem conditions and
\(C_t=(c_{t,1},\ldots,c_{t,n_{\scriptstyle t}})\)
the ordered key-claim sequence. Suppose all key claims pass reliable local verification; that is, for every \(i\in\{1,\ldots,n_t\}\),
\begin{equation}
\label{eq:agpv-conditional-local-soundness}
P_q\land\bigwedge_{j=1}^{i-1}c_{t,j}\models c_{t,i},
\end{equation}
where \(\models\) denotes semantic entailment. For \(i=1\), Eq.~\eqref{eq:agpv-conditional-local-soundness} yields \(P_q\models c_{t,1}\); induction over the ordered claims then gives
\begin{equation}
\label{eq:agpv-conditional-claim-chain}
P_q\models\bigwedge_{i=1}^{n_{\scriptstyle t}}c_{t,i}.
\end{equation}
Further suppose the problem objective contains completeness requirements
\(h_1,\ldots,h_m\), and every \(h_\ell\) is semantically entailed by some key claim
\(c_{t,j(\ell)}\). Equation~\eqref{eq:agpv-conditional-claim-chain} and transitivity then yield
\begin{equation}
\label{eq:agpv-conditional-completeness}
P_q\models\bigwedge_{\ell=1}^{m}h_\ell.
\end{equation}
Thus, if the key claims sufficiently cover final-answer correctness and completeness and local verification is reliable, a single process-verification chain supports all these requirements. Complete definitions and proofs are provided in Appendix~\ref{app:conditional_soundness}.

\section{Experiments}

Our experiments address three questions: (1) whether AGPV improves final-answer accuracy; (2) whether its gains primarily arise from correcting initially incorrect answers rather than from indiscriminate rewriting; and (3) whether its performance remains stable across backbone models and settings with different initial capabilities.

\begin{table}[!t]
\centering
\small
\setlength{\tabcolsep}{5pt}
\renewcommand{\arraystretch}{1.12}
\begin{tabularx}{\linewidth}{l*{5}{>{\centering\arraybackslash}X}}
\toprule
Method & \shortstack{DeepSeek\\V4.1-Flash} & \shortstack{DeepSeek\\V4-Pro (0813)} & \shortstack{Qwen3.6\\35B-A3B-FP8} & \shortstack{GPT-5.6\\Luna} & Mean \\
\midrule
Initial & 52.9 & 38.8 & 53.5 & 23.5 & 42.2 \\
Reflex & \accgain{60.0}{\scoreup{+7.1}} & \accgain{37.6}{\scoredown{-1.2}} & \accgain{54.7}{\scoreup{+1.2}} & \accgain{31.2}{\scoreup{+7.6}} & \accgain{45.9}{\scoreup{+3.7}} \\
Self-Refine & \accgain{61.8}{\scoreup{+8.8}} & \accgain{51.2}{\scoreup{+12.4}} & \accgain{57.6}{\scoreup{+4.1}} & \accgain{30.6}{\scoreup{+7.1}} & \accgain{50.3}{\scoreup{+8.1}} \\
Self-Reflection & \accgain{55.3}{\scoreup{+2.4}} & \accgain{40.0}{\scoreup{+1.2}} & \accgain{57.1}{\scoreup{+3.5}} & \accgain{32.9}{\scoreup{+9.4}} & \accgain{46.3}{\scoreup{+4.1}} \\
CheckList & \accgain{55.3}{\scoreup{+2.4}} & \accgain{41.2}{\scoreup{+2.4}} & \accgain{57.1}{\scoreup{+3.5}} & \accgain{30.0}{\scoreup{+6.5}} & \accgain{45.9}{\scoreup{+3.7}} \\
CoT-Tool & \accgain{54.1}{\scoreup{+1.2}} & \accgain{40.0}{\scoreup{+1.2}} & \accgain{41.8}{\scoredown{-11.8}} & \accgain{22.4}{\scoredown{-1.2}} & \accgain{39.6}{\scoredown{-2.6}} \\
ProgCo & \accgain{58.2}{\scoreup{+5.3}} & \accgain{41.2}{\scoreup{+2.4}} & \accgain{57.1}{\scoreup{+3.5}} & \accgain{25.9}{\scoreup{+2.4}} & \accgain{45.6}{\scoreup{+3.4}} \\
ProgCo-Py & \accgain{62.4}{\scoreup{+9.4}} & \accgain{48.2}{\scoreup{+9.4}} & \accgain{58.8}{\scoreup{+5.3}} & \accgain{28.2}{\scoreup{+4.7}} & \accgain{49.4}{\scoreup{+7.2}} \\
\textbf{AGPV} & \accgain{\textbf{72.4}}{\scoreup{+19.4}} & \accgain{\textbf{66.5}}{\scoreup{+27.6}} & \accgain{\textbf{64.1}}{\scoreup{+10.6}} & \accgain{\textbf{52.4}}{\scoreup{+28.8}} & \accgain{\textbf{63.8}}{\scoreup{+21.6}} \\
\bottomrule
\end{tabularx}
\caption{Final-answer accuracy (\%). Model scores pool 170 problems across five datasets; Mean averages four models (680 model--problem instances). Subscripts give changes from Initial in percentage points, computed from unrounded counts (green/red: increases/decreases). Bold marks column bests; per-dataset results appear in the appendix.}
\label{tab:main_results_non_thinking}
\end{table}

\subsection{Experimental Setup}

\paragraph{Datasets}
We evaluate the methods on five hard-test mathematical reasoning datasets comprising 170 problems: AIME 2024~\citepboxed{huggingfaceh42025aime2024}, AIME 2025~\citepboxed{yentinglin2025aime}, the MathArena releases of BRUMO 2025 and HMMT February 2025~\citepboxed{balunovic2025matharena}, and AMO Bench~\citepboxed{an2025amobench}. These datasets cover algebra, combinatorics, number theory, geometry, and integrated reasoning in competition mathematics. Dataset sizes and basic characteristics are provided in Table~\ref{tab:benchmark_overview}.
\textbf{Backbone Models.}
The experiments use DeepSeek-V4.1-Flash and DeepSeek-V4-Pro (0813)~\citepboxed{deepseek2026api}, Qwen3.6-35B-A3B-FP8~\citepboxed{qwen2026model}, and GPT-5.6-Luna~\citepboxed{openai2026luna}. Short model names follow the retained experiment records; the complete mapping is given in Table~\ref{tab:model_decoding_configuration}.

\paragraph{Baselines}
We compare AGPV with three representative classes of baselines. Initial Score denotes the backbone model's initial response without additional correction. The first class comprises natural-language correction methods: Reflex, which uses natural-language reflection; Self-Refine~\citepboxed{madaan2023selfrefine}, which uses iterative self-feedback; Self-Reflection, which uses reflective feedback; and CheckList~\citepboxed{cook2024ticking}, which performs checklist-guided correction. The second class is the reasoning-enhanced method CoT-Tool~\citepboxed{abedi_cogitator_2025}, which samples multiple chain-of-thought paths~\citepboxed{wei2022chain} and selects an answer using Self-Consistency~\citepboxed{wang2022selfconsistency}. The third class comprises the verification-enhanced methods ProgCo~\citepboxed{song-etal-2025-progco} and ProgCo-Py. ProgCo uses program-driven verification and correction, whereas ProgCo-Py enables Python-tool feedback based on the public ProgCo implementation. AGPV instead performs process verification along the key-claim sequence of the current solution and uses the first critical error and tool evidence to guide subsequent correction.

\paragraph{Experimental Protocol}
All self-correction methods use \(\texttt{max\_turn}=3\) and follow their respective stopping rules. This setting does not imply matched model-call or token budgets. All self-correction methods start from the same pregenerated responses. We analyze intermediate answers retrospectively from these runs rather than from independent \(\texttt{max\_turn}=1\) experiments. Empty outputs or outputs without a valid extractable answer are incorrect; when execution falls back to a saved answer, that returned answer is evaluated. Accuracy always uses the original fixed benchmark denominator.
\textbf{Implementation Parameters.}
The generation temperature is set to \(0.0\) for all methods. Following its own mechanism, CoT-Tool independently generates three chain-of-thought paths and determines its output by the majority among the extracted answers; its budget is therefore not interpreted as correction turns. Complete model identifiers, reasoning modes, method-level budgets, tool-execution limits, and answer-judgment rules are provided in Appendix~\ref{app:experiment_reproducibility}.
\textbf{Evaluation Metrics.}
Final-answer accuracy is our primary metric. Each model's overall score is computed over all 170 problems, weighting datasets by their sample counts. Mean averages these four model scores, equivalently pooling 680 model--problem instances, not 680 distinct questions. For correction analysis, \(C\) and \(W\) denote correct and incorrect answers. Thus, \(W\rightarrow C\) denotes an initially incorrect answer becoming correct and \(C\rightarrow W\) denotes an initially correct answer becoming incorrect. We additionally report the average number of recorded states as a description of the saved processes, not a measure of total inference cost.

\subsection{Main Results}

AGPV leads all evaluated baselines under each of the four models (\tabref{tab:main_results_non_thinking}). Mean accuracy reaches \(63.8\%\), exceeding Self-Refine, the strongest evaluated baseline, by \(13.5\) percentage points. Gains span models with different initial capabilities, indicating that the benefit is not restricted to one backbone. Additional robustness analyses are provided in Appendix~\ref{app:complete_results}.

\noindent\textbf{Per-Dataset Results.}
Tables~\ref{tab:appendix_deepseek_results} and \ref{tab:appendix_qwen_gpt_results} retain the complete breakdown over the five datasets. Overall improvements are consistent across the four models, although AGPV does not lead every individual dataset. We next examine the sources and distribution of these gains.

\subsection{Correction-Behavior Analysis}

\begin{figure}[!t]
\centering
\begin{minipage}[t]{0.485\textwidth}
\vspace{0pt}
\centering
\includegraphics[width=0.95\linewidth]{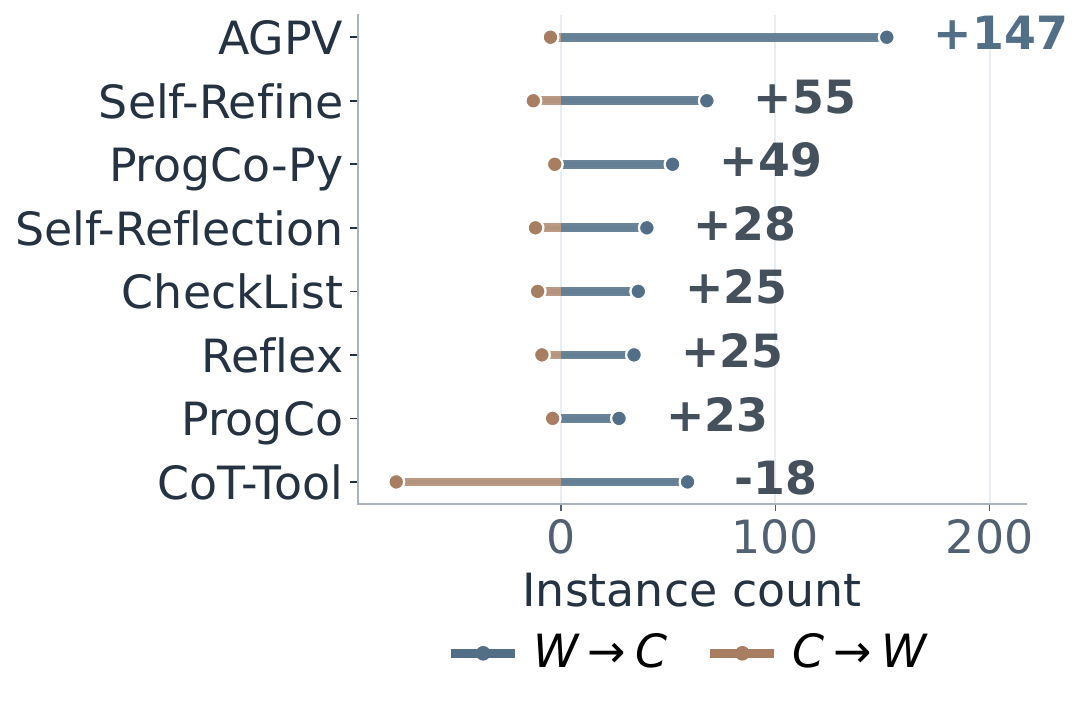}
\captionof{figure}{Initial-to-final repair and damage counts. Blue/right: \(W\rightarrow C\) repairs; orange/left: \(C\rightarrow W\) damage. Labels and descending order use net repairs, \(N_{W\rightarrow C}-N_{C\rightarrow W}\).}
\label{fig:repair_benefit_damage_cost}
\end{minipage}\hfill
\begin{minipage}[t]{0.485\textwidth}
\vspace{0pt}
\centering
\includegraphics[width=\linewidth]{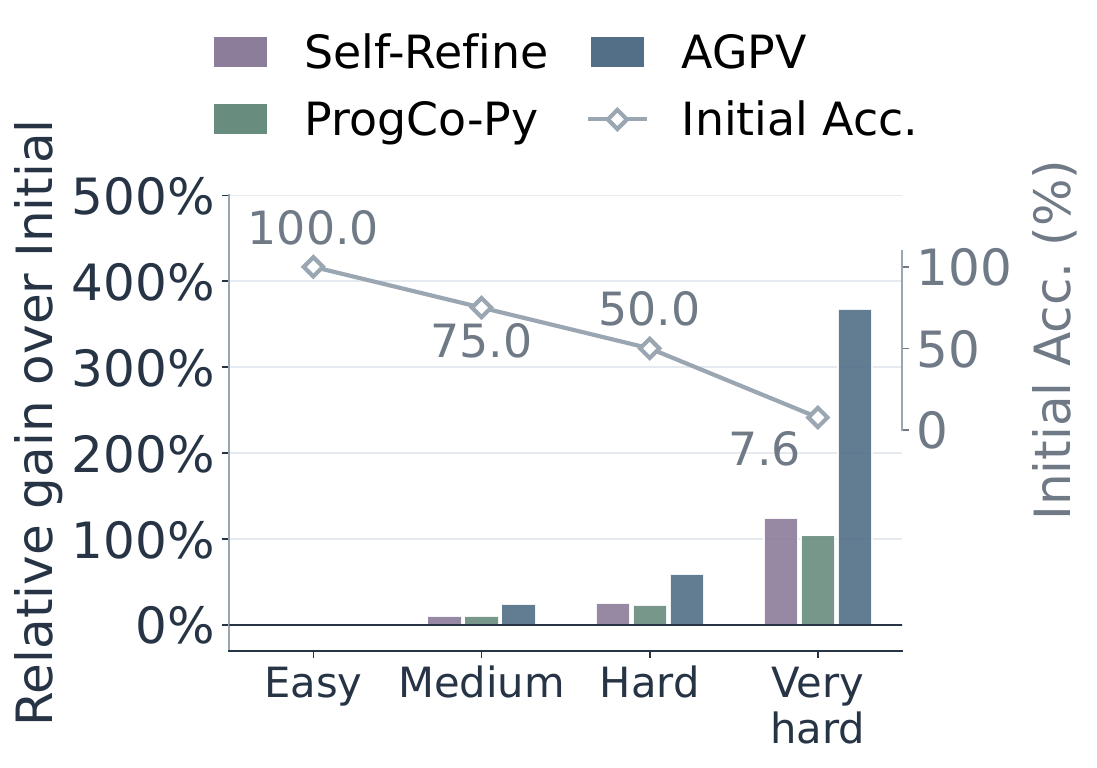}
\captionof{figure}{Relative accuracy gains (bars, left axis) and initial accuracy (gray line, right axis) across empirical-difficulty groups.}
\label{fig:difficulty_turn_accuracy}
\end{minipage}
\end{figure}

To identify the source of AGPV's gains, we further measure each method's transitions between initial and final correctness. The main text focuses on \(W\rightarrow C\) repairs and \(C\rightarrow W\) damage; complete state transitions and derived statistics are reported in Table~\ref{tab:correction_transition_summary}.
\figref{fig:repair_benefit_damage_cost} shows that AGPV's gains primarily come from broader error repair: it corrects 152 initially incorrect answers while damaging 5 initially correct answers. Repair benefits substantially outweigh these regressions, producing the largest net improvement among the evaluated methods.

\begin{figure}[!t]
\centering
\begin{minipage}[t]{0.485\textwidth}
\vspace{0pt}
\centering
\begingroup
\small
\renewcommand{\arraystretch}{1.12}
\begin{tabularx}{\linewidth}{@{}Xr@{}}
\toprule
Repairing method(s) & Share (\%) \\
\midrule
\multicolumn{2}{@{}l}{\textit{Disjoint repair groups}} \\
AGPV only & 62.7 \\
AGPV and ProgCo-Py & 15.3 \\
AGPV and ProgCo & 2.8 \\
All three methods & 5.1 \\
\midrule
ProgCo-Py only & 6.8 \\
ProgCo only & 5.1 \\
ProgCo and ProgCo-Py & 2.3 \\
\midrule
\multicolumn{2}{@{}l}{\textit{Total coverage (overlapping)}} \\
ProgCo & 15.3 \\
ProgCo-Py & 29.4 \\
AGPV & 85.9 \\
\bottomrule
\end{tabularx}
\par
\endgroup
\captionof{table}{Repair coverage (\%) among 177 model--problem instances repaired by at least one method. Upper groups are exclusive: ``and'' means only the named, independently run methods repair an instance, not joint execution. Lower totals overlap. Percentages are rounded.}
\label{tab:w2c_repair_overlap}
\end{minipage}\hfill
\begin{minipage}[t]{0.485\textwidth}
\vspace{0pt}
\centering
\includegraphics[width=\linewidth]{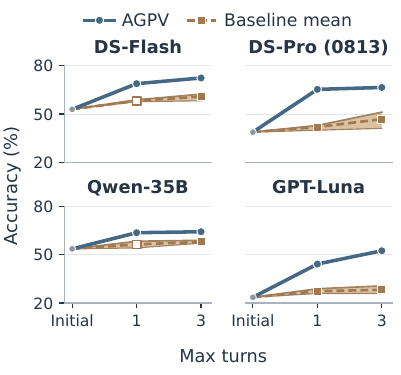}
\captionof{figure}{Accuracy at reconstructed stopping points. The dashed line and shaded band show the mean and min--max range of ProgCo, ProgCo-Py, and Self-Refine, respectively. Hollow markers indicate means containing imputed values.}
\label{fig:accuracy_turn_proxy}
\end{minipage}
\end{figure}

\tabref{tab:w2c_repair_overlap} compares the disjoint overlap of repairs by ProgCo, ProgCo-Py, and AGPV. Each combination identifies methods that repair the same initially incorrect instances. AGPV alone repairs 111 instances left unrepaired by both baselines, showing that its advantage extends beyond their shared repair coverage.

\par\begingroup\noindent\ignorespaces
Table~\ref{tab:pairwise_repair_by_model} in Appendix~\ref{app:zh:stopping} compares AGPV with each of four baselines on the same initially incorrect answers, problem by problem, for each of the four models.
\par\endgroup

\subsection{Performance by Difficulty}

To examine whether AGPV's gains concentrate in particular difficulty ranges, we estimate empirical difficulty from initial-answer correctness across multiple backbone-model settings. Specifically, a problem is considered empirically more difficult when fewer of the four models answer it correctly at initialization. We label problems solved initially by all four models as Easy, by three models as Medium, by two models as Hard, and by zero or one model as Very hard. These groups contain 30, 26, 32, and 82 distinct problems, respectively.

\figref{fig:difficulty_turn_accuracy} compares Self-Refine, ProgCo-Py, and AGPV across these groups. Relative gain is defined as \((\mathrm{Acc}_{\mathrm{Final}}-\mathrm{Acc}_{\mathrm{Initial}})/\mathrm{Acc}_{\mathrm{Initial}}\). Because low initial accuracy magnifies this ratio, the figure also shows Initial Acc.

AGPV exceeds both baselines on Medium, Hard, and Very hard problems, with an absolute gain of \(28.0\) percentage points over Initial in the Very hard group. Its advantage therefore concentrates in problems that the initial models solve less reliably. Easy problems exhibit a small regression under AGPV, while both baselines preserve their initial accuracy.

\subsection{Effect of Iteration Turns}

AGPV verifies and corrects the current solution through feedback. To examine how accuracy changes during this process, \figref{fig:accuracy_turn_proxy} compares AGPV with ProgCo, ProgCo-Py, and Self-Refine across all four models. We reconstruct stopping-point answers for limits of one and three turns using each method's stopping rule, retaining an earlier terminal answer after stopping. Missing intermediate scores are completed with the same answer-evaluation procedure, without regenerating responses; unavailable intermediate answers use the recorded fallback. Reconstructed stopping-point accuracies are provided in Table~\ref{tab:saved_trajectory_counts}.

Across all four models, AGPV exceeds the best of the three baselines at both reported post-initial checkpoints, not only their mean. Most gains are already present at the first reported stopping point, with further improvements by the final point. Table~\ref{tab:correction_transition_summary} describes recorded-state counts; neither these counts nor saved checkpoints imply equal computation budgets. Appendix~\ref{app:context_compression} reports the observed reduction in request-context length when compression is triggered, rather than an overall token-efficiency comparison.

\paragraph{Computational Cost.}
\par\begingroup\noindent\ignorespaces
On DeepSeek Flash, AGPV achieved 72.4\% accuracy using an average of 89.23k tokens per problem, and Self-Refine achieved 61.8\% using 35.95k (Table~\ref{tab:agpv_added_cost}, Appendix~\ref{app:recorded_token_cost}). Table~\ref{tab:agpv_added_budget} and Figure~\ref{fig:agpv_token_budget} in Appendix~\ref{app:zh:budget} compare three methods under the same token budgets.
\par\endgroup

\section{Conclusion}

AGPV verifies ordered key claims, locates the first failed claim, and uses failure evidence to revise subsequent reasoning and update the verification plan. Embedding completeness requirements in these claims unifies correctness and completeness checks; both are conditionally guaranteed with sufficient claim coverage, reliable local verification, and all claims passing. Experiments demonstrate improved mathematical-reasoning accuracy, error correction, and repair coverage. Future work will extend process verification to other domains with traceable intermediate processes, including code generation, scientific reasoning, and data analysis, using domain knowledge and specialized tools to construct well-scoped claims verifiable with reliable evidence.

\section*{AI Use Statement}
Generative AI tools were used for language polishing, literature retrieval, discussions of research design, drafting portions of the experimental descriptions, and assistance with code and figure preparation. The core research ideas were developed by the authors, who reviewed the final content and take responsibility for the paper.

\bibliographystyle{iclr2027_conference}
\bibliography{references}

\clearpage
\setlength{\textfloatsep}{\agpvsavedtextfloatsep}
\setlength{\floatsep}{\agpvsavedfloatsep}
\setlength{\intextsep}{\agpvsavedintextsep}
\setlength{\abovecaptionskip}{\agpvsavedabovecaptionskip}
\appendix
\setcounter{secnumdepth}{2}
\begin{center}
{\LARGE\bfseries Appendix}
\end{center}
\section{Overall AGPV Algorithm}
\label{app:overall_algorithm}

\begin{algorithm}[H]
\caption{Overall AGPV Procedure}
\label{alg:agfv}
\begin{algorithmic}[1]
\REQUIRE Problem \(q\), initial solution \(y_0\), extracted answer \(a_0\), maximum zero-based iteration index \(T\)
\ENSURE Final solution \(y\) and answer \(a\)
\STATE Initialize \(y \leftarrow y_0,\ a \leftarrow a_0,\ C\leftarrow\varnothing,\ V \leftarrow \varnothing\)
\FOR{\(t=0\) to \(T\)}
    \IF{\(V=\varnothing\)}
        \STATE \((C,V)\leftarrow \mathcal{A}_{\mathrm{gen}}(q,y,a)\) \COMMENT{Generate the key-claim sequence and its verification plan}
    \ENDIF
    \STATE \((r,R,k)\leftarrow \mathcal{A}_{\mathrm{ver}}(q,y,a,C,V)\) \COMMENT{Verify key claims; invoke \(\mathcal{A}_{\mathrm{tool}}\) as needed; locate the first critical error}
    \IF{\(r=1\)}
        \STATE \RETURN \(y,a\) \COMMENT{Verification passes; accept the current answer}
    \ENDIF
    \STATE \(F\leftarrow \mathcal{A}_{\mathrm{sum}}(q,y,a,V,R,k)\) \COMMENT{Summarize failure evidence}
    \STATE \(y'\leftarrow \mathcal{A}_{\mathrm{fix}}(q,y,a,F)\) \COMMENT{Correct from the first critical error; invoke \(\mathcal{A}_{\mathrm{tool}}\) as needed}
    \STATE \(a'\leftarrow \operatorname{Extract}(y')\) \COMMENT{Extract the corrected answer}
    \IF{\(y'=\varnothing\ \lor\ a'=\varnothing\)}
        \STATE \RETURN \(y,a\) \COMMENT{No valid correction generated; retain the current candidate}
    \ENDIF
    \IF{\(I_{\mathrm{ans}}(a',a)=0\)}
        \STATE \(y \leftarrow y',\ a \leftarrow a',\ C\leftarrow\varnothing,\ V \leftarrow \varnothing\) \COMMENT{Answers differ; regenerate \(C,V\) next turn}
    \ELSE
        \STATE \((C,V)\leftarrow \mathcal{A}_{\mathrm{upd}}(q,y',a',C,V,R,F)\) \COMMENT{Answers are equivalent; update \(C,V\); invoke tools as needed}
        \STATE \(y \leftarrow y',\ a \leftarrow a'\)
    \ENDIF
\ENDFOR
\STATE \RETURN \(y,a\) \COMMENT{Iteration limit reached; return the latest candidate without further verification}
\end{algorithmic}
\end{algorithm}
\par\begingroup\noindent\ignorespaces
Table~\ref{tab:agpv_added_prompts} lists the core instructions for each agent in Figure~\ref{fig:agfv_framework}. The supplementary file \texttt{AGPV\_agent\_prompts\_en.md} contains the full instructions, input templates, and examples, organized by agent.

\noindent\begin{minipage}{\linewidth}
\captionof{table}{Prompts for the agents in Figure~\ref{fig:agfv_framework}}\label{tab:agpv_added_prompts}
{\small\setlength{\tabcolsep}{4pt}\renewcommand{\arraystretch}{1.08}
\begin{tabularx}{\linewidth}{@{}p{.27\linewidth}X@{}}\toprule
Agent & Core instruction (abridged) \\ \midrule
$\mathcal A_{\mathrm{gen}}$ Generate key claims and a verification plan & Extract the key claims from the current solution and plan the checks in reasoning order. Include the problem's requirements for a complete answer. \\
$\mathcal A_{\mathrm{ver}}$ Verify key claims & Check each claim according to the plan, using the mathematical toolbox as needed. Stop at the first failed claim and record the evidence. \\
$\mathcal A_{\mathrm{sum}}$ Summarize evidence of failure & Use the verification record to identify the failed step, explain why it failed, and indicate where correction should begin. \\
$\mathcal A_{\mathrm{fix}}$ Revise the solution & Use the feedback to retain the valid reasoning, redo the derivation from the error onward, and provide a complete solution. \\
$\mathcal A_{\mathrm{upd}}$ Update the key claims and verification plan & When the answer stays the same after correction, recheck the original plan against the revised solution and verification record. Retain applicable checks, add missing ones, and adjust those that no longer match the reasoning. \\
$\mathcal A_{\mathrm{tool}}$ Shared mathematical toolbox & Turn mathematical requests from the other agents into executable Python code, perform symbolic solving, enumeration, or constraint checks, and return the results. \\
\bottomrule\end{tabularx}\par}
\end{minipage}\par
\par\endgroup

\section{Theoretical Supplement}

\subsection{Conditional Reliability Proof for Process Verification}
\label{app:conditional_soundness}

To formalize how completeness is embedded in process verification, we denote the completeness requirements in the problem objective by
\begin{equation}
\label{eq:app-completeness-requirements}
\mathcal{H}(q,y_t)=\{h_1,h_2,\ldots,h_m\},
\end{equation}
Here, \(\mathcal{H}(q,y_t)\) is the set of completeness requirements induced by the problem objective and instantiated under the current solution \(y_t\); \(\ell\in\{1,\ldots,m\}\) indexes these requirements, and \(m\) is their total number. Each \(h_\ell\) denotes a completeness property or proposition required by the problem objective rather than the target object itself. For example, \(h_\ell\) may state that “the target object given by the current solution equals the complete target object implied by the problem conditions.” The key-claim sequence \(C_t\) covers these completeness requirements if, for every \(h_\ell\in\mathcal{H}(q,y_t)\), there exists a key claim \(c_{t,j(\ell)}\), where \(j(\ell)\in\{1,\ldots,n_t\}\) is the position of the key claim covering the \(\ell\)-th completeness requirement, such that the claim semantically entails the corresponding requirement:
\begin{equation}
\label{eq:app-completeness-coverage}
c_{t,j(\ell)}\models h_\ell .
\end{equation}
Here, \(\models\) denotes semantic entailment, and \(c_{t,j(\ell)}\models h_\ell\) states that the \(j(\ell)\)-th key claim semantically entails the corresponding completeness requirement. This definition embeds completeness into the process-verification chain when the key-claim sequence is constructed.

\phantomsection\label{app:example_completeness_x2}
\begin{agfvexamplebox}
\noindent\textbf{Example: Instantiating a Completeness Requirement.}

For the problem \(x^2=4\), suppose the current solution gives only \(x=2\). Checking only whether \(2\) satisfies the equation is insufficient. The problem objective requires the complete set of real solutions, and solving under the problem conditions yields the complete target object
\begin{equation}
\label{eq:app-solution-set-x2}
S^*=\{x\in\mathbb{R}:x^2=4\}=\{-2,2\}.
\end{equation}
If \(S_y\) is the set returned by the current solution, the corresponding completeness requirement can be written as
\begin{equation}
\label{eq:app-completeness-requirement-x2}
h:\quad S_y=S^*.
\end{equation}
AGPV constructs a key claim associated with this objective, such as whether the solution set returned by the current solution is the complete real solution set. The set returned by the current solution is
\begin{equation}
\label{eq:app-incomplete-solution-set-x2}
S_y=\{2\}.
\end{equation}
Since
\begin{equation}
\label{eq:app-set-mismatch-x2}
S_y\neq S^*,
\end{equation}
the key claim fails verification, allowing AGPV to detect the missing solution.
\end{agfvexamplebox}

\paragraph{Conditional-Reliability Intuition}
AGPV's process verification admits an idealized conditional-reliability interpretation. To simplify notation, we omit the turn subscript \(t\) below and denote the current solution by \(y\). Let \(P_q\) be the problem conditions and let the key-claim sequence be
\begin{equation}
\label{eq:app-claim-sequence-no-turn}
C=(c_1,c_2,\ldots,c_n).
\end{equation}
where \(n\) is the total number of key claims after omitting the turn subscript. For \(i\in\{1,\ldots,n\}\), define the verified context preceding the \(i\)-th claim as
\begin{equation}
\label{eq:app-verified-context}
\Gamma_{i-1}=P_q\land c_1\land c_2\land\cdots\land c_{i-1},
\end{equation}
where \(\Gamma_0=P_q\). Ideally, if the verifier is reliable, then whenever the \(i\)-th key claim passes verification,
\begin{equation}
\label{eq:app-local-soundness}
\Gamma_{i-1}\models c_i.
\end{equation}
Here, \(\models\) denotes semantic entailment. The relation \(\Gamma_{i-1}\models c_i\) means that the current key claim \(c_i\) holds in every interpretation satisfying the problem conditions and the preceding verified claims. Thus, the \(i\)-th key claim follows from the problem conditions and the preceding verified claims.

\paragraph{Proposition 1: Conditional Reliability of Process Verification}
Let \(P_q\) be the problem conditions and let \(C=(c_1,\ldots,c_n)\) be the ordered key-claim sequence constructed by AGPV from the current solution and the problem objective. Suppose that all key claims pass verification and that the verifier is reliable; that is, for every \(i\),
\begin{equation}
\label{eq:app-proposition-assumption}
P_q\land c_1\land\cdots\land c_{i-1}\models c_i,
\end{equation}
where the set of preceding claims is empty when \(i=1\). Then
\begin{equation}
\label{eq:app-proposition-conclusion}
P_q\models c_1\land c_2\land\cdots\land c_n.
\end{equation}

\paragraph{Proof}
We proceed by induction on \(i\).

Base case: when \(i=1\), verifier reliability gives
\begin{equation}
\label{eq:app-base-step}
P_q\models c_1.
\end{equation}
Therefore, the proposition holds for \(i=1\).

Induction hypothesis: suppose that, for some \(i-1\geq 1\),
\begin{equation}
\label{eq:app-induction-hypothesis}
P_q\models c_1\land c_2\land\cdots\land c_{i-1}.
\end{equation}
That is, the first \(i-1\) key claims are all supported by the problem conditions \(P_q\).

Inductive step: because the \(i\)-th key claim passes verification and the verifier is reliable,
\begin{equation}
\label{eq:app-induction-local-step}
P_q\land c_1\land\cdots\land c_{i-1}\models c_i.
\end{equation}
From the induction hypothesis and \(P_q\models P_q\), conjunction introduction gives \(P_q\models\Gamma_{i-1}\). Because \(\Gamma_{i-1}\models c_i\), transitivity of semantic entailment yields \(P_q\models c_i\). Applying conjunction introduction again to this conclusion and the induction hypothesis gives
\begin{equation}
\label{eq:app-induction-next}
P_q\models c_1\land\cdots\land c_i.
\end{equation}
Mathematical induction therefore establishes Eq.~\eqref{eq:app-proposition-conclusion}. This completes the proof.

\paragraph{Corollary 1: Completeness Is Covered by the Same Process-Verification Chain}
Suppose the construction of the key-claim sequence \(C\) covers the completeness requirements \(\mathcal{H}(q,y)=\{h_1,\ldots,h_m\}\) associated with the current solution; that is, for every \(h_\ell\in\mathcal{H}(q,y)\), there exists a key claim \(c_{j(\ell)}\) such that
\begin{equation}
\label{eq:app-corollary-coverage}
c_{j(\ell)}\models h_\ell,
\end{equation}
Then, under the conditions of Proposition 1,
\begin{equation}
\label{eq:app-corollary-conclusion}
P_q\models h_1\land h_2\land\cdots\land h_m.
\end{equation}

\paragraph{Proof}
Proposition 1 gives \(P_q\models c_1\land\cdots\land c_n\), and hence \(P_q\models c_{j(\ell)}\) for every \(j(\ell)\). Combining this result with Eq.~\eqref{eq:app-corollary-coverage}, transitivity of semantic entailment yields \(P_q\models h_\ell\). This conclusion holds for every \(h_\ell\in\mathcal{H}(q,y)\); conjunction introduction then gives Eq.~\eqref{eq:app-corollary-conclusion}. This completes the proof.

The proposition and corollary clarify the meaning of embedded completeness in AGPV. When a problem requires all solutions, complete counting, exhaustive enumeration, or a global optimum, these completeness requirements enter the same process-verification chain through the construction of the key-claim sequence. If the key claims provide complete coverage and the corresponding verification is reliable, the completeness conclusions are also supported by the problem conditions.

This conditional-reliability intuition is consistent with soundness theorems in formal logic: if a conclusion follows from a set of premises through sound inference rules, then the conclusion is a semantic consequence of those premises. Soundness proofs for natural deduction commonly proceed by induction over derivations~\citepboxed{openlogic_soundness}. The same idea is also related to Floyd--Hoare-style program verification, in which intermediate assertions organize local checks and composition rules support global properties~\citepboxed{floyd1967assigning,hoare1969axiomatic,softwarefoundations_hoare}.

AGPV is not a formal proof checker. The proposition and corollary characterize how process verification can embed completeness requirements under the ideal conditions that the key claims provide complete coverage, their order matches the original reasoning process, and local verification is reliable. Under these conditions, the first key claim that fails verification corresponds to the first critical error in the current solution. In practice, performance may still be affected by incomplete key-claim extraction, completeness requirements omitted from the key-claim sequence, incorrectly constructed verification targets, or tool-execution failures.

\section{Supplementary Experimental Results}

This appendix provides additional information for the experiments in the main paper, including experimental settings and reproduction protocols, a benchmark overview, complete model results, supplementary correction-behavior analyses, and descriptive statistics of recorded states. The main paper reports the primary results most relevant to the central narrative, the core correction-behavior analysis, and the difficulty-conditioned analysis; this appendix documents the full configurations, complete numerical results, complete model-setting results, and additional statistical analyses.

\subsection{Experimental Settings and Reproduction Protocol}
\label{app:experiment_reproducibility}

To ensure that results are comparable and reproducible across methods and model settings, this section reports the experimental configuration from four perspectives: model and decoding configurations, method budgets, AGPV tool-execution limits, and unified answer evaluation. Unless otherwise specified, all methods under the same model setting use the same model interface and reasoning mode, with the generation temperature set to \(0.0\). Provider-specific parameters that are not uniformly overridden by the experimental scripts retain the configurations of their respective API wrappers; interface parameters with different semantics are not forcibly aligned.

\paragraph{Model and Decoding Configurations}
\tabref{tab:model_decoding_configuration} maps the four model settings to their API identifiers or model checkpoint. All four use the non-thinking wrapper configuration. The label 0813 identifies the retained DeepSeek-Pro experiment tag; it is not an additional model setting. The same model-specific initial responses are reused for every correction method.

\begin{table}[H]
\centering
\small
\renewcommand{\arraystretch}{1.10}
\setlength{\tabcolsep}{3pt}
\begin{tabularx}{\linewidth}{@{}l>{\raggedright\arraybackslash}X>{\raggedright\arraybackslash}X@{}}
\toprule
Model setting & API identifier / checkpoint & Reasoning configuration \\
\midrule
DeepSeek-V4.1-Flash & \texttt{deepseek-flash} & Thinking disabled \\
DeepSeek-V4-Pro (0813) & \texttt{deepseek-v4-pro} & Thinking disabled \\
Qwen3.6-35B-A3B-FP8 & \texttt{Qwen/\allowbreak Qwen3.6-\allowbreak 35B-A3B-FP8} & \texttt{enable\_thinking=\allowbreak False} \\
GPT-5.6-Luna & \texttt{gpt-5.6-luna} & \texttt{reasoning\_effort=\allowbreak none} \\
\bottomrule
\end{tabularx}
\caption{Model identifiers and decoding configurations for the retained four-model experiments.}
\label{tab:model_decoding_configuration}
\end{table}

\paragraph{Method Budgets and Fair Comparison}
\tabref{tab:method_budget_configuration} summarizes the initial-response source and execution budget for each method. ProgCo, ProgCo-Py, Reflex, Self-Refine, Self-Reflection, CheckList, and AGPV start from the same pregenerated responses. CoT-Tool instead generates three independent chain-of-thought paths. All retained correction runs use \(\texttt{max\_turn}=3\) under method-specific stopping rules. The shared setting does not imply identical model calls, token usage, or runtime. For example, Self-Refine generates feedback and a revision within one turn, whereas CheckList generates a checklist before correction.

The iteration analysis uses saved intermediate answers from these retained runs; it does not compare independent \(\texttt{max\_turn}\in\{1,2,3\}\) experiments or perform a parameter sweep. Generation temperature is fixed at \(0\) when exposed by the API. Other method-specific settings are retained rather than forcibly aligned across different execution mechanisms.

\begin{table}[H]
\centering
\small
\renewcommand{\arraystretch}{1.12}
\setlength{\tabcolsep}{5pt}
\begin{tabularx}{\linewidth}{@{}>{\raggedright\arraybackslash}p{0.17\linewidth}>{\raggedright\arraybackslash}p{0.17\linewidth}>{\raggedright\arraybackslash}X@{}}
\toprule
Method & Initial response & Main-experiment budget and execution mechanism \\
\midrule
ProgCo & Shared & Up to three correction turns; program-driven verification and correction. \\
ProgCo-Py & Shared & Up to three correction turns; Python-tool feedback enabled within the ProgCo workflow. \\
Reflex & Shared & Up to three correction turns; one reflection-and-rewrite step per turn. \\
Self-Refine & Shared & Up to three correction turns; feedback is generated before each feedback-guided revision. \\
Self-Reflection & Shared & Up to three correction turns; one self-reflective correction per turn. \\
CheckList & Shared & One checklist is generated first, followed by up to three checklist-guided correction turns. \\
CoT-Tool & Independently generated & Independently generates three chain-of-thought paths and selects the output by majority vote over the extracted answers. \\
AGPV & Shared & \(\texttt{max\_turn}=3\); locates the first failed key claim, generates feedback, and regenerates or adapts the verification plan after correction. \\
\bottomrule
\end{tabularx}
\caption{Main-experiment budgets and execution mechanisms for all methods.}
\label{tab:method_budget_configuration}
\end{table}

For reproducibility, every correction method receives the same stored initial response and candidate answer under a given model--problem setting. Reported accuracies use one retained run per model--method--problem combination. Existing answers and final-answer scores are preserved; only missing intermediate-answer judgments are completed for the retrospective trajectory analysis.

\paragraph{AGPV Tool Execution and Runtime Stability}
AGPV's verification, correction, and verification-plan update agents can invoke the shared mathematical toolbox dynamically. The backends support symbolic simplification, equation solving, enumeration, constraint checking, and exact or numerical computation. The four retained AGPV runs set a language-model-call timeout of 960 seconds and a Python execution timeout of 120 seconds; ordinary and stopping-decision calls have retry limits of 10 and 20, respectively. A soft warning and hard stop are configured at 100 and 120 generated Python snippets per sample. The working-context compression trigger is 131,072 tokens, subject to the runtime's model-capacity checks. Successful nonempty results are cached by problem identifier and trimmed tool-request text; cache hits return the previous evidence instead of generating and executing Python again. These are implementation settings, not an ablation or a measured token-saving result. Failed executions and their returned fallback answers remain in the fixed-denominator evaluation.

\paragraph{Unified Answer Evaluation}
All methods use the same three-stage answer-evaluation pipeline: normalized exact matching, directly decidable numerical comparison, and a model-based equivalence judgment when needed. Parsed numerical answers are considered equivalent when their absolute error is below \(10^{-3}\). Expressions, sets, and other unresolved answer forms are judged with the shared DeepSeek-Flash fallback at temperature \(0.0\). Set-valued answers must contain all required elements, and order is checked when the problem requires an ordered answer. Empty or unextractable answers are incorrect. If execution returns an earlier saved answer after a failure, we score that actual returned answer; no sample is removed. The fixed denominators are 30 each for AIME 2024, AIME 2025, BRUMO 2025, and HMMT February 2025, and 50 for AMO Bench, totaling 170 distinct problems per model.

\paragraph{Code Availability}
Code is available at \url{https://github.com/TicusFFF/AGPV}.

\subsection{Benchmark Overview}

\tabref{tab:benchmark_overview} summarizes the five hard-test mathematical reasoning datasets used in this work. AIME 2024~\citepboxed{huggingfaceh42025aime2024}, AIME 2025~\citepboxed{yentinglin2025aime}, and the MathArena releases of BRUMO 2025 and HMMT February 2025~\citepboxed{balunovic2025matharena} each contain 30 problems; AMO Bench~\citepboxed{an2025amobench} contains 50 problems. Together, the five datasets comprise 170 problems spanning algebra, number theory, combinatorics, geometry, and integrated reasoning in competition mathematics.

\begin{table}[htbp]
\centering
\small
\renewcommand{\arraystretch}{1.12}
\begin{tabularx}{\columnwidth}{lcrX}
\toprule
Dataset & Short & \#Problems & Notes \\
\midrule
AIME 2024 & A24 & 30 & competition math; integer answers \\
AIME 2025 & A25 & 30 & competition math; integer answers \\
BRUMO 2025 & B25 & 30 & hard contest-style reasoning \\
HMMT Feb. 2025 & HMM & 30 & contest math; mixed topics \\
AMO Bench & AMO & 50 & advanced math reasoning; mixed answer forms \\
\bottomrule
\end{tabularx}
\caption{Overview of the five hard-test mathematical reasoning datasets.}
\label{tab:benchmark_overview}
\end{table}

\subsection{Complete Results}
\label{app:complete_results}

Tables~\ref{tab:appendix_deepseek_results} and \ref{tab:appendix_qwen_gpt_results} give the full five-dataset breakdown for the same four models as the main table. All scores use fixed denominators, including empty outputs and recorded failure fallbacks. AGPV has the highest overall accuracy for each model, but it is not the best method on every individual dataset; for example, Self-Refine is stronger on HMMT under DeepSeek-V4.1-Flash.

\begin{table}[H]
\centering
\small
\setlength{\tabcolsep}{3pt}
\renewcommand{\arraystretch}{1.12}
\begin{tabular}{lrrrrrrrrrrrr}
\toprule
& \multicolumn{6}{c}{DeepSeek-V4.1-Flash}  & \multicolumn{6}{c}{DeepSeek-V4-Pro (0813)} \\
\cmidrule(lr){2-7}\cmidrule(lr){8-13}
Method & A24 & A25 & B25 & HMM & AMO & Avg & A24 & A25 & B25 & HMM & AMO & Avg \\
\midrule
Initial & 73.3 & 73.3 & 86.7 & 53.3 & 8.0 & 52.9 & 63.3 & 46.7 & 66.7 & 26.7 & 10.0 & 38.8 \\
Reflex & 83.3 & 80.0 & 90.0 & 60.0 & 16.0 & \accgain{60.0}{\scoreup{+7.1}} & 63.3 & 43.3 & 63.3 & 30.0 & 8.0 & \accgain{37.6}{\scoredown{-1.2}} \\
Self-Refine & \textbf{86.7} & 83.3 & 83.3 & \textbf{70.0} & 16.0 & \accgain{61.8}{\scoreup{+8.8}} & 70.0 & 60.0 & 86.7 & 56.7 & 10.0 & \accgain{51.2}{\scoreup{+12.4}} \\
Self-Reflection & 80.0 & 70.0 & 90.0 & 60.0 & 8.0 & \accgain{55.3}{\scoreup{+2.4}} & \textbf{76.7} & 46.7 & 66.7 & 26.7 & 6.0 & \accgain{40.0}{\scoreup{+1.2}} \\
CheckList & 76.7 & 76.7 & 86.7 & 53.3 & 12.0 & \accgain{55.3}{\scoreup{+2.4}} & 66.7 & 46.7 & 66.7 & 33.3 & 12.0 & \accgain{41.2}{\scoreup{+2.4}} \\
CoT-Tool & 83.3 & 73.3 & 76.7 & 56.7 & 10.0 & \accgain{54.1}{\scoreup{+1.2}} & 66.7 & 46.7 & 66.7 & 30.0 & 10.0 & \accgain{40.0}{\scoreup{+1.2}} \\
ProgCo & \textbf{86.7} & 73.3 & 86.7 & 60.0 & 14.0 & \accgain{58.2}{\scoreup{+5.3}} & 70.0 & 50.0 & 70.0 & 26.7 & 10.0 & \accgain{41.2}{\scoreup{+2.4}} \\
ProgCo-Py & 83.3 & 76.7 & 90.0 & 63.3 & 24.0 & \accgain{62.4}{\scoreup{+9.4}} & 70.0 & 56.7 & 83.3 & 43.3 & 12.0 & \accgain{48.2}{\scoreup{+9.4}} \\
\textbf{AGPV} & 83.3 & \textbf{96.7} & \textbf{96.7} & 60.0 & \textbf{44.0} & \accgain{\textbf{72.4}}{\scoreup{+19.4}} & 73.3 & \textbf{80.0} & \textbf{93.3} & \textbf{76.7} & \textbf{32.0} & \accgain{\textbf{66.5}}{\scoreup{+27.6}} \\
\bottomrule
\end{tabular}
\caption{Per-dataset accuracy (\%) for DeepSeek-V4.1-Flash and DeepSeek-V4-Pro (0813). Avg is sample-weighted over 170 problems. Changes beside Avg are percentage-point differences from the shared Initial score, calculated from unrounded counts. Bold marks the best result in each column, including ties.}
\label{tab:appendix_deepseek_results}
\end{table}

\begin{table}[H]
\centering
\small
\setlength{\tabcolsep}{3pt}
\renewcommand{\arraystretch}{1.12}
\begin{tabular}{lrrrrrrrrrrrr}
\toprule
& \multicolumn{6}{c}{Qwen3.6-35B-A3B-FP8} & \multicolumn{6}{c}{GPT-5.6-Luna} \\
\cmidrule(lr){2-7}\cmidrule(lr){8-13}
Method & A24 & A25 & B25 & HMM & AMO & Avg & A24 & A25 & B25 & HMM & AMO & Avg \\
\midrule
Initial & 73.3 & 60.0 & 73.3 & 56.7 & 24.0 & 53.5 & 40.0 & 36.7 & 40.0 & 16.7 & 0.0 & 23.5 \\
Reflex & 70.0 & 60.0 & 73.3 & 66.7 & 24.0 & \accgain{54.7}{\scoreup{+1.2}} & 56.7 & 40.0 & 56.7 & 20.0 & 2.0 & \accgain{31.2}{\scoreup{+7.6}} \\
Self-Refine & \textbf{80.0} & 66.7 & 73.3 & 63.3 & 26.0 & \accgain{57.6}{\scoreup{+4.1}} & 53.3 & 43.3 & 46.7 & 23.3 & 4.0 & \accgain{30.6}{\scoreup{+7.1}} \\
Self-Reflection & 73.3 & 60.0 & 76.7 & 63.3 & 30.0 & \accgain{57.1}{\scoreup{+3.5}} & 56.7 & 40.0 & 53.3 & 30.0 & 4.0 & \accgain{32.9}{\scoreup{+9.4}} \\
CheckList & \textbf{80.0} & 63.3 & 73.3 & 63.3 & 26.0 & \accgain{57.1}{\scoreup{+3.5}} & 53.3 & 40.0 & 43.3 & 23.3 & 6.0 & \accgain{30.0}{\scoreup{+6.5}} \\
CoT-Tool & 60.0 & 56.7 & 56.7 & 43.3 & 12.0 & \accgain{41.8}{\scoredown{-11.8}} & 50.0 & 26.7 & 33.3 & 16.7 & 0.0 & \accgain{22.4}{\scoredown{-1.2}} \\
ProgCo & \textbf{80.0} & 70.0 & 73.3 & 60.0 & 24.0 & \accgain{57.1}{\scoreup{+3.5}} & 53.3 & 36.7 & 43.3 & 13.3 & 0.0 & \accgain{25.9}{\scoreup{+2.4}} \\
ProgCo-Py & \textbf{80.0} & \textbf{73.3} & 76.7 & 60.0 & 26.0 & \accgain{58.8}{\scoreup{+5.3}} & 46.7 & 40.0 & 46.7 & 23.3 & 2.0 & \accgain{28.2}{\scoreup{+4.7}} \\
\textbf{AGPV} & \textbf{80.0} & 70.0 & \textbf{80.0} & \textbf{70.0} & \textbf{38.0} & \accgain{\textbf{64.1}}{\scoreup{+10.6}} & \textbf{66.7} & \textbf{66.7} & \textbf{80.0} & \textbf{50.0} & \textbf{20.0} & \accgain{\textbf{52.4}}{\scoreup{+28.8}} \\
\bottomrule
\end{tabular}
\caption{Per-dataset accuracy (\%) for Qwen3.6-35B-A3B-FP8 and GPT-5.6-Luna. Avg is sample-weighted over 170 problems. Subscripts show percentage-point changes from the shared Initial score, calculated from unrounded counts. Bold marks the best result in each column, including ties.}
\label{tab:appendix_qwen_gpt_results}
\end{table}

\par\begingroup\noindent\ignorespaces
Initial accuracy of Qwen and DeepSeek Flash\par\nobreak
Tables~\ref{tab:appendix_deepseek_results} and~\ref{tab:appendix_qwen_gpt_results} report the initial-answer results for the two models. We compared the models' initial answers using the same test problems and scoring procedure. Overall accuracy is computed across all test problems, with equal weight per problem. Qwen and DeepSeek Flash each led on two datasets and tied on the fifth. Qwen's initial-answer accuracy on AMO Bench was 16.0 percentage points higher. In problem counts, Qwen's extra correct answers on AMO Bench exactly offset DeepSeek Flash's combined lead on its two stronger datasets. The overall gap comes from HMMT 2025, leaving Qwen ahead by 0.6 percentage points.
\par\endgroup

\paragraph{Robustness Analysis}
Using the retained runs, we compute 95\% paired bootstrap confidence intervals with 50,000 resamples stratified by dataset, resampling problems jointly across the four models. We also recompute accuracy gains after excluding each dataset in turn.

\begin{table}[!htbp]
\centering
\small
\setlength{\tabcolsep}{9pt}
\renewcommand{\arraystretch}{1.12}
\begin{tabular}{lrrr}
\toprule
Baseline & Mean gain & 95\% CI & \shortstack{Minimum gain after\\excluding one dataset} \\
\midrule
Self-Refine & +13.5 & [9.9, 17.4] & +11.0 \\
ProgCo-Py & +14.4 & [10.9, 18.1] & +13.1 \\
ProgCo & +18.2 & [14.6, 22.1] & +16.9 \\
\bottomrule
\end{tabular}
\caption{Accuracy gains of AGPV over three baselines, in percentage points. Intervals quantify problem-sampling uncertainty conditional on the retained runs; minimum gains are taken over the five leave-one-dataset-out evaluations.}
\label{tab:robustness_analysis}
\end{table}

AGPV retains a positive average advantage over all three baselines after excluding any single dataset.

\subsection{Supplementary Correction-Behavior Analysis}
\label{app:zh:stopping}

\tabref{tab:correction_transition_summary} summarizes the complete correction-state transitions and associated repair, damage, final-accuracy, and recorded-state statistics for all methods.

\begin{table}[!htbp]
\centering
\small
\setlength{\tabcolsep}{1.3pt}
\renewcommand{\arraystretch}{1.12}
\begin{minipage}[t]{0.49\linewidth}
\vspace{0pt}
\small
\centering
\textbf{State transitions (\%)}\\[2pt]
\begin{tabular}{@{}lrrrr@{}}
\toprule
\shortstack{~\\Method} & $C\rightarrow C$ & $C\rightarrow W$ & $W\rightarrow C$ & $W\rightarrow W$ \\
\midrule
Reflex & 40.9 & 1.3 & 5.0 & 52.8 \\
Self-Refine & 40.3 & 1.9 & 10.0 & 47.8 \\
\shortstack[l]{Self-\\Reflection} & 40.4 & 1.8 & 5.9 & 51.9 \\
CheckList & 40.6 & 1.6 & 5.3 & 52.5 \\
CoT-Tool & 30.9 & 11.3 & 8.7 & 49.1 \\
ProgCo & 41.6 & 0.6 & 4.0 & 53.8 \\
ProgCo-Py & 41.8 & 0.4 & 7.6 & 50.1 \\
\textbf{AGPV} & 41.5 & 0.7 & 22.4 & 35.4 \\
\bottomrule
\end{tabular}
\end{minipage}\hfill
\begin{minipage}[t]{0.49\linewidth}
\vspace{0pt}
\small
\centering
\textbf{Repair, damage, and outcome statistics}\\[2pt]
\begin{tabular}{@{}lrrrrr@{}}
\toprule
Method & \shortstack{Repair\\(\%)} & \shortstack{Damage\\(\%)} & \shortstack{Net gain\\(pp)} & \shortstack{Final\\Acc. (\%)} & \shortstack{Avg\\States} \\
\midrule
Reflex & 8.7 & 3.1 & +3.7 & 45.9 & 3.9 \\
Self-Refine & 17.3 & 4.5 & +8.1 & 50.3 & 3.8 \\
\shortstack[l]{Self-\\Reflection} & 10.2 & 4.2 & +4.1 & 46.3 & 3.9 \\
CheckList & 9.2 & 3.8 & +3.7 & 45.9 & 3.9 \\
CoT-Tool & 15.0 & 26.8 & -2.6 & 39.6 & 1.0 \\
ProgCo & 6.9 & 1.4 & +3.4 & 45.6 & 1.8 \\
ProgCo-Py & 13.2 & 1.0 & +7.2 & 49.4 & 1.9 \\
\textbf{AGPV} & 38.7 & 1.7 & +21.6 & 63.8 & 1.9 \\
\bottomrule
\end{tabular}
\end{minipage}
\caption{Correction-state transitions and repair/damage statistics. State transitions are percentages of 680 model--problem instances; \(C\) and \(W\) denote correct and incorrect answers, and \(X\rightarrow Y\) denotes the initial-to-final transition. Repair and Damage use initially incorrect and initially correct samples as denominators, respectively. Net gain is final minus initial accuracy in percentage points (pp), computed before rounding. Avg States is the mean number of saved \texttt{turns} entries, not a computation budget.}
\label{tab:correction_transition_summary}
\end{table}

AGPV achieves substantially broader repair coverage than ProgCo and ProgCo-Py, which accounts for most of its net gain despite a slightly higher damage count. For CoT-Tool, transitions compare its independently generated final answer with the shared Initial reference; they do not describe an iterative repair procedure.

Recorded-state conventions differ by method. Natural-language baselines generally include the initial state and subsequent revisions; verification methods log attempted verification states, and a final corrected answer need not create another verified-state entry. CoT-Tool records an aggregate result rather than one state per generation path. These counts do not normalize model calls, tokens, tool calls, or runtime and therefore do not establish an efficiency advantage.

\par\begingroup\noindent\ignorespaces
Table~\ref{tab:agpv_added_stopping} groups the 680 outputs from the four models by how verification ended. The second column gives each group's share of all outputs, and the third gives accuracy within that group.

\noindent\begin{minipage}{\linewidth}
\captionof{table}{Final accuracy by return condition}\label{tab:agpv_added_stopping}
{\small\setlength{\tabcolsep}{4pt}\renewcommand{\arraystretch}{1.08}
\begin{tabularx}{\linewidth}{@{}Xrr@{}}\toprule
Return condition & Share of outputs & Accuracy within group \\ \midrule
Original solution returned after passing the first check & 52.1\% & 75.4\% \\
Passed a later check after failing the first & 35.0\% & 62.6\% \\
Returned at the round limit after failing the last check & 12.9\% & 20.5\% \\
\bottomrule\end{tabularx}\par}
\end{minipage}\par
Accuracy was highest for solutions that passed the first check, followed by those that passed after correction. Of the 88 solutions that reached the round limit and failed the last check, 18 had correct final answers (20.5\%). Incorrect final answers in this group accounted for 10.3\% of all outputs.
\par\endgroup

\par\begingroup\noindent\ignorespaces
Table~\ref{tab:agpv_added_repair} shows each method's correction rate on four models: the fraction of initially incorrect answers that were correct at the end. Each row pools one model's results across five datasets. All methods started from the same initial answers.

\noindent\begin{minipage}{\linewidth}
\captionof{table}{Correction coverage across methods}\label{tab:agpv_added_repair}
{\small\setlength{\tabcolsep}{4pt}\renewcommand{\arraystretch}{1.08}
\begin{tabularx}{\linewidth}{@{}Xrrrrr@{}}\toprule
Model & AGPV & ProgCo & ProgCo-Py & Self-Refine & Self-Reflection \\ \midrule
DeepSeek Flash & 42.5\% & 11.3\% & 20.0\% & 26.3\% & 11.3\% \\
DeepSeek Pro & 48.1\% & 6.7\% & 17.3\% & 26.0\% & 5.8\% \\
Qwen 35B & 22.8\% & 7.6\% & 11.4\% & 8.9\% & 10.1\% \\
GPT-5.6 Luna & 38.5\% & 3.8\% & 6.9\% & 10.0\% & 13.1\% \\
\bottomrule\end{tabularx}\par}
\end{minipage}\par
Across the 393 initially incorrect answers from the four models, AGPV corrected 38.7\%, Self-Refine 17.3\%, and Self-Reflection 10.2\%. AGPV also corrected 69.1\% (47/68) of the answers corrected by Self-Refine and 62.5\% (25/40) of those corrected by Self-Reflection. For each model and baseline, Table~\ref{tab:pairwise_repair_by_model} reports the proportions corrected by both methods, by only one, and by neither.
\par\endgroup

\par\begingroup\noindent\ignorespaces
\begin{table}[H]
\centering
\caption[Pairwise repair outcomes by model and baseline]{Pairwise correction outcomes for AGPV and each baseline on the same initially incorrect answers}\label{tab:pairwise_repair_by_model}
\smallskip
{\small\setlength{\tabcolsep}{2pt}\renewcommand{\arraystretch}{1.08}
\begin{tabular*}{\linewidth}{@{\extracolsep{\fill}}llrrrr@{}}
\toprule
Model & Baseline & \shortstack[r]{AGPV\\only} & \shortstack[r]{Baseline\\only} & \shortstack[r]{Both\\corrected it} & \shortstack[r]{Neither\\corrected it} \\
\midrule
DeepSeek Flash & ProgCo & 36.3\% & 5.0\% & 6.3\% & 52.5\% \\
 & ProgCo-Py & 27.5\% & 5.0\% & 15.0\% & 52.5\% \\
 & Self-Refine & 23.8\% & 7.5\% & 18.8\% & 50.0\% \\
 & Self-Reflection & 35.0\% & 3.8\% & 7.5\% & 53.8\% \\
\addlinespace[3pt]
DeepSeek Pro & ProgCo & 43.3\% & 1.9\% & 4.8\% & 50.0\% \\
 & ProgCo-Py & 33.7\% & 2.9\% & 14.4\% & 49.0\% \\
 & Self-Refine & 28.8\% & 6.7\% & 19.2\% & 45.2\% \\
 & Self-Reflection & 45.2\% & 2.9\% & 2.9\% & 49.0\% \\
\addlinespace[3pt]
Qwen 35B & ProgCo & 20.3\% & 5.1\% & 2.5\% & 72.2\% \\
 & ProgCo-Py & 17.7\% & 6.3\% & 5.1\% & 70.9\% \\
 & Self-Refine & 16.5\% & 2.5\% & 6.3\% & 74.7\% \\
 & Self-Reflection & 17.7\% & 5.1\% & 5.1\% & 72.2\% \\
\addlinespace[3pt]
GPT-5.6 Luna & ProgCo & 36.9\% & 2.3\% & 1.5\% & 59.2\% \\
 & ProgCo-Py & 34.6\% & 3.1\% & 3.8\% & 58.5\% \\
 & Self-Refine & 33.1\% & 4.6\% & 5.4\% & 56.9\% \\
 & Self-Reflection & 29.2\% & 3.8\% & 9.2\% & 57.7\% \\
\bottomrule
\end{tabular*}\par}
\smallskip
\begin{minipage}{\linewidth}\small
Each row compares AGPV and one baseline on the same initially incorrect answers. The denominator for all four percentages is the total number of initially incorrect answers for that model. Percentages may not sum to 100\% because of rounding.
\end{minipage}
\end{table}
\par\endgroup

\FloatBarrier
\Needspace{9\baselineskip}
\subsection{Computational Cost and Context Compression}
\label{app:recorded_token_cost}

\par\begingroup\noindent\ignorespaces
AGPV and the iterative correction baselines start from the same initial answers. Table~\ref{tab:agpv_added_cost} reports accuracy and mean cost per problem on five datasets with DeepSeek Flash. Model calls and Python executions are counts per problem. Token counts are in thousands per problem. We estimated ProgCo-Py's token usage from run logs.

\noindent\begin{minipage}{\linewidth}
\captionof{table}{Accuracy and mean cost per problem by method}\label{tab:agpv_added_cost}
{\footnotesize\setlength{\tabcolsep}{2.7pt}\renewcommand{\arraystretch}{1.08}
\begin{tabularx}{\linewidth}{@{}Xrrrrrrr@{}}\toprule
Method & Accuracy & \shortstack{Model\\calls} & \shortstack{Input\\tokens} & \shortstack{Output\\tokens} & \shortstack{Total\\tokens} & \shortstack{Python\\runs} & \shortstack{Time\\(min)} \\ \midrule
Shared initial answers & 52.9\% & 1.86 & 2.97 & 3.54 & 6.51 & 0 & 0.57 \\
Reflex & 60.0\% & 3.00 & 10.33 & 9.04 & 19.37 & 0 & 0.61 \\
Self-Refine & 61.8\% & 6.00 & 23.77 & 12.18 & 35.95 & 0 & 0.92 \\
Self-Reflection & 55.3\% & 3.00 & 11.55 & 10.66 & 22.21 & 0 & 0.77 \\
CheckList & 55.3\% & 4.00 & 11.22 & 8.74 & 19.96 & 0 & 0.65 \\
CoT-Tool & 54.1\% & 3.00 & 0.58 & 10.29 & 10.87 & 0 & 0.72 \\
ProgCo & 58.2\% & 7.82 & 45.31 & 11.93 & 57.24 & 0 & 1.07 \\
ProgCo-Py & 62.4\% & 48.55 & 16,945.88 & 58.61 & 17,004.49 & 39.79 & 31.76 \\
AGPV & 72.4\% & 14.71 & 80.33 & 8.90 & 89.23 & 4.06 & 2.07 \\
\bottomrule\end{tabularx}\par}
\end{minipage}\par
AGPV outperformed Self-Refine by 10.6 percentage points. The additional input was mainly used for stepwise verification: the model reads the reasoning steps and tool outputs before revising the solution. Compared with ProgCo-Py, which also uses Python, AGPV achieved 10.0 percentage points higher accuracy with nearly 90\% fewer Python executions per problem. When the same tool request recurs within a problem, AGPV reuses its previous successful result, avoiding 17 duplicate Python executions in this experiment.
\par\endgroup

\paragraph{Recorded AGPV Component Token Costs}
Table~\ref{tab:recorded_token_cost} reports backend-returned token usage for AGPV's tool-code generation and successful context-compression calls. Values are averaged over all 170 problems per model, including problems that do not trigger either component. Tool-code generation is counted once from the timing records, without adding the duplicate entries stored in tool-execution records; compression usage is counted once per recorded successful event.

\begin{table}[htbp]
\centering
\small
\setlength{\tabcolsep}{6pt}
\renewcommand{\arraystretch}{1.12}
\begin{tabular}{lrrrr}
\toprule
& \multicolumn{2}{c}{Tool-code generation} & \multicolumn{2}{c}{Context compression} \\
\cmidrule(lr){2-3}\cmidrule(lr){4-5}
Backbone & Input & Output & Input & Output \\
\midrule
DS-Flash & 8.08 & 0.96 & 0.00 & 0.00 \\
DS-Pro (0813) & 21.54 & 2.31 & 15.50 & 0.08 \\
Qwen-35B & 3.33 & 0.39 & 3.54 & 0.01 \\
GPT-Luna & 6.27 & 0.59 & 0.00 & 0.00 \\
\bottomrule
\end{tabular}
\caption{Recorded AGPV token-cost components, in thousands of tokens per problem. Input and output use the respective backends' reported token counts. Zeros for Flash and Luna compression mean that no compression was triggered. These are component costs, not total inference costs.}
\label{tab:recorded_token_cost}
\end{table}

This accounting excludes other verification, correction, plan-generation, summarization, and answer-extraction calls, shared initial-answer generation, evaluation, and any unrecorded attempts. The retained ProgCo-Py and Self-Refine outputs do not provide matching token-usage records; their missing costs are not treated as zero. Thus, the table quantifies observable AGPV overhead but does not support a total-cost ranking against these baselines. Compression itself consumes tokens, so the request-length reductions below cannot be interpreted as net token savings.

\paragraph{Python Execution Requests}
Table~\ref{tab:python_execution_requests} separately compares Python execution requests in the retained AGPV and ProgCo-Py runs. These counts measure tool usage, not total inference cost.

\begin{table}[htbp]
\centering
\small
\setlength{\tabcolsep}{6pt}
\renewcommand{\arraystretch}{1.12}
\begin{tabular}{lrrrr}
\toprule
Method & DS-Flash & DS-Pro (0813) & Qwen-35B & GPT-Luna \\
\midrule
AGPV & 4.065 & 11.035 & 1.494 & 3.206 \\
ProgCo-Py & 39.788 & 10.471 & 3.124 & 0.006 \\
\bottomrule
\end{tabular}
\caption{Mean Python execution requests per problem, averaged over all 170 problems per model. Requests resulting in execution errors or timeouts are included; startup checks and cache hits are excluded.}
\label{tab:python_execution_requests}
\end{table}

\paragraph{Request-Context Compression}
\label{app:context_compression}

Figure~\ref{fig:context_compression} compares request-context lengths immediately before and after compression over all recorded compression events in the retained runs. The DeepSeek-Pro (0813) run records 22 events across 6 problems, and the Qwen run records 5 events across 5 problems. DeepSeek-Flash and GPT-Luna are not shown because compression was not triggered in their retained runs. Repeated events on the same problem are included. Bars show event-level mean lengths, and the reduction ratio is the total length reduction divided by the total pre-compression length.

\par\noindent
\begin{minipage}[t]{0.485\textwidth}
\vspace{0pt}
\centering
\includegraphics[width=\linewidth]{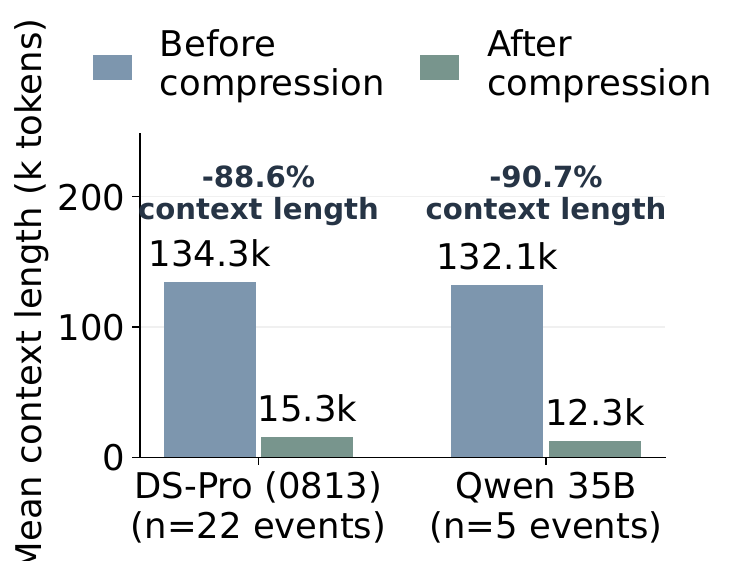}
\captionof{figure}{Request-context lengths before and after compression, measured in thousands of tokens. All recorded compression events are included; \(n\) denotes events, not distinct problems. These reductions measure request length, not total inference-token savings.}
\label{fig:context_compression}
\end{minipage}\hfill
\begin{minipage}[t]{0.485\textwidth}
\vspace{0pt}
Compression reduces aggregate request-context length by \(88.6\%\) for DeepSeek-Pro (0813) and \(90.7\%\) for Qwen, leaving more compact inputs for continued verification. These observations characterize the executed compression operations, not an accuracy ablation or a total-cost comparison with ProgCo-Py or Self-Refine. Complete per-call token usage is not available across these methods, and compression itself incurs model-call cost; therefore, these context reductions should not be interpreted as net inference-token savings.
\end{minipage}
\par\medskip

\clearpage
\begingroup
\setlength{\intextsep}{4pt}
\setlength{\abovecaptionskip}{4pt}
\subsection{Saved-Checkpoint Accuracy}
\label{app:zh:budget}

\begin{table}[H]
\centering
\small
\setlength{\tabcolsep}{4pt}
\renewcommand{\arraystretch}{1.12}
\begin{tabular}{llrrrr}
\toprule
Model & Method & Initial & \(\texttt{max\_turn}=1\) & \(\texttt{max\_turn}=2\) & \(\texttt{max\_turn}=3\) \\
\midrule
DS-Flash & AGPV & 52.9 & 68.8 & 70.0 & 72.4 \\
DS-Flash & ProgCo & 52.9 & 57.6 & 57.6 & 58.2 \\
DS-Flash & ProgCo-Py & 52.9 & 58.8 & 62.4 & 62.4 \\
DS-Flash & Self-Refine & 52.9 & 58.2 & 60.6 & 61.8 \\
\midrule
DS-Pro (0813) & AGPV & 38.8 & 65.3 & 66.5 & 66.5 \\
DS-Pro (0813) & ProgCo & 38.8 & 40.0 & 41.2 & 41.2 \\
DS-Pro (0813) & ProgCo-Py & 38.8 & 42.9 & 45.9 & 48.2 \\
DS-Pro (0813) & Self-Refine & 38.8 & 42.9 & 45.9 & 51.2 \\
\midrule
Qwen-35B & AGPV & 53.5 & 63.5 & 64.1 & 64.1 \\
Qwen-35B & ProgCo & 53.5 & 56.5 & 57.1 & 57.1 \\
Qwen-35B & ProgCo-Py & 53.5 & 58.2 & 58.8 & 58.8 \\
Qwen-35B & Self-Refine & 53.5 & 54.1 & 56.5 & 57.6 \\
\midrule
GPT-Luna & AGPV & 23.5 & 44.1 & 50.0 & 52.4 \\
GPT-Luna & ProgCo & 23.5 & 25.9 & 25.3 & 25.9 \\
GPT-Luna & ProgCo-Py & 23.5 & 27.1 & 28.2 & 28.2 \\
GPT-Luna & Self-Refine & 23.5 & 28.8 & 30.6 & 30.6 \\
\bottomrule
\end{tabular}
\caption{Accuracy (\%) on 170 problems per model at initialization and reconstructed stopping points (\texttt{max\_turn}=1, 2, 3).}
\label{tab:saved_trajectory_counts}
\end{table}

\par\begingroup\noindent\ignorespaces
We compared three methods on five datasets under the same token budgets. All methods started from the same initial answers. The budget counts only the input and output tokens used after the initial answer. For each recorded DeepSeek Flash trajectory, we took the last complete answer within the budget, using the initial answer if no correction had been completed. For Table~\ref{tab:agpv_added_budget}, we rescored these answers using the same three-step procedure. Table~\ref{tab:main_results_non_thinking} reports the original scores.

\noindent\begin{minipage}{\linewidth}
\captionof{table}{Accuracy under the same additional token budgets}\label{tab:agpv_added_budget}
{\small\setlength{\tabcolsep}{4pt}\renewcommand{\arraystretch}{1.08}
\begin{tabularx}{\linewidth}{@{}Xrrr@{}}\toprule
Additional token limit per problem & AGPV & Self-Refine & ProgCo-Py \\ \midrule
40,000 & 54.7\% & 59.4\% & 52.9\% \\
90,000 & 65.9\% & 61.8\% & 55.9\% \\
700,000 & 74.1\% & 62.4\% & 62.9\% \\
\bottomrule\end{tabularx}\par}
\end{minipage}\par
At 40,000 tokens per problem, Self-Refine had the highest accuracy. As Figure~\ref{fig:agpv_token_budget} shows, AGPV overtook Self-Refine at about 61,000 tokens. As the budget rose from 90,000 to 700,000 tokens, AGPV's accuracy increased from 65.9\% to 74.1\%, the largest gain over this interval.
\begin{center}\begin{minipage}{\linewidth}\centering
\includegraphics[width=\linewidth,height=.245\textheight,keepaspectratio]{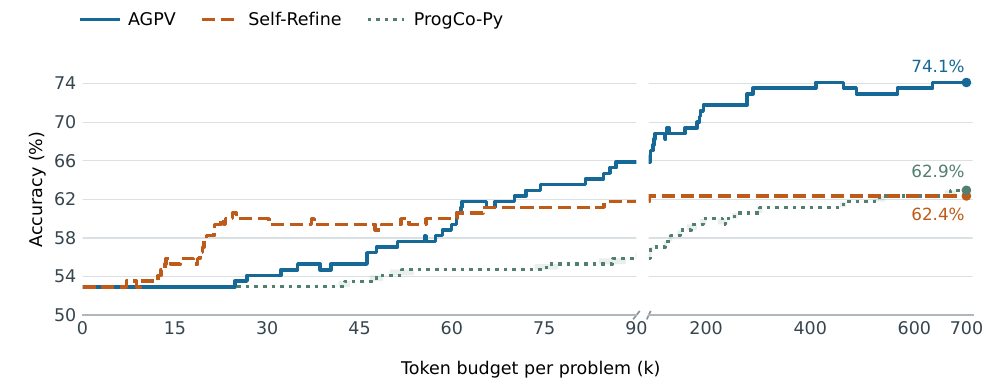}\par
\captionof{figure}{Accuracy by additional token budget per problem. We estimated ProgCo-Py's token usage from run logs.}\label{fig:agpv_token_budget}
\end{minipage}\end{center}
\par\endgroup

\subsection{Comparative Case Analysis}

\tabref{tab:case_diagnosis} presents four illustrative successful corrections from the retained four-model runs, with model names and sample identifiers stated explicitly. The table summarizes the saved diagnoses and answers; it is not a frequency estimate of error types or a formal proof of every generated derivation.

\begin{table}[H]
\centering
\small
\setlength{\tabcolsep}{3.5pt}
\renewcommand{\arraystretch}{1.28}
\arrayrulecolor{black!45}
\resizebox{\linewidth}{!}{%
\begin{minipage}{7in}
\centering
\begin{tabular}{p{0.09\textwidth}|p{0.155\textwidth}p{0.17\textwidth}|p{0.155\textwidth}p{0.22\textwidth}p{0.105\textwidth}}
\toprule
Dataset & Problem-objective summary & Initial error & ProgCo / ProgCo-Py behavior & AGPV key evidence and correction & Characteristic \\
\midrule
HMMT February 2025
& Count circular arrangements of \(1,\ldots,7\) with adjacent products at most \(20\); rotations and reflections are distinct.
& DeepSeek-Pro (0813), \texttt{hmmt\_feb\_2025\_10}: the initial answer is \(28\), not \(56\). After fixing \(7\), the counting argument wrongly excludes arrangements where \(6\) is adjacent to \(1\).
& Both methods retain \(28\).
& Enumeration returns \(56\) valid ordered arrangements and confirms that \(7\)'s neighbors are \(1,2\). Feedback identifies the incomplete count; the corrected argument includes the omitted placements and returns \(56\).
& Counting-coverage diagnosis and correction. \\
\midrule
AIME 2024
& Compute a product over the thirteenth roots of unity and return its remainder modulo \(1000\).
& GPT-Luna, \texttt{aime\_2024\_19}: the initial answer is \(65\), not \(321\). The factorization is valid, but the sign of \((1+i)^{13}\) is wrong.
& Both methods retain \(65\).
& Verification identifies \((1+i)^{13}=-64-64i\), rather than \(64+64i\). Recomputing the conjugate product gives \(8321\), hence remainder \(321\).
& Local arithmetic-error diagnosis. \\
\midrule
AIME 2025
& Find the expected number of regions formed by two perpendicular diameters and \(25\) random cross-quadrant chords.
& DeepSeek-Flash, \texttt{aime\_2025\_12}: the initial answer is \(487/3\), not \(204\). It assumes random chords intersect with probability \(1/3\), ignoring the quadrant-conditioned sampling.
& Both methods retain \(487/3\).
& The saved feedback gives the conditioned probability \(17/36\); the correction trace checks it numerically and recomputes the expected intersections and regions, returning \(204\).
& Intermediate-probability correction. \\
\midrule
AMO Bench
& Find all possible remainders modulo \(32\) of \(\lfloor n!/((n+1)(n+2))\rfloor\).
& DeepSeek-Pro (0813), \texttt{amo\_bench\_6}: the initial set is \(\{0,2,4,6,12,18\}\), omitting valid values.
& ProgCo returns a set containing an extra \(21\); ProgCo-Py returns an incomplete seven-element set.
& The saved verification enumerates \(1\leq n\leq1000\) and finds missing even values. AGPV returns the reference-matching set \(\{0,2,4,\ldots,30\}\), explicitly enumerated in the saved answer.
& Missing-value detection from finite witnesses. \\
\bottomrule
\end{tabular}
\arrayrulecolor{black}
\end{minipage}%
}
\caption{Illustrative corrections from the current four-model runs. Problem identifiers follow the stored benchmark files. In the AMO case, finite enumeration supplies witnesses of omitted values; it is not by itself a proof of exhaustiveness for all \(n\).}
\label{tab:case_diagnosis}
\end{table}
\endgroup

\par\begingroup\noindent\ignorespaces
Detecting and recovering missing solutions\par\nobreak
We examined 12 initial answers from 4 models on 3 AMO Bench problems. Seven listed specific integers but omitted some valid values. Of these 7 initial answers, AGPV identified specific missing values in 5 during the first round and ultimately produced complete, correct answers in 4.
\par
The residue problem in Table~\ref{tab:case_diagnosis} asks for all possible residues of $\left\lfloor n!/((n+1)(n+2))\right\rfloor$ modulo 32 as $n$ ranges over the positive integers. DeepSeek Pro's original answer listed 6 residues. AGPV found the 10 missing ones, and its final list of 16 matched the reference answer. Qwen 35B's original answer listed 7 residues. In the first round, AGPV enumerated $n=1$ to $50$ in Python and added the missing residues 14, 20, 24, and 30. In the second round, the verifier enumerated the same range. The 11 residues it found matched the revised answer, which passed verification. Yet $n=51$ already gives 28, which was missing from the answer. Verification correctly computed the residues within the chosen range but did not establish that the range covered every possible residue. Compared with the reference answer, AGPV still missed 8, 10, 18, 22, and 28.
\par\endgroup

\par\begingroup\noindent\ignorespaces
Using a counterexample to locate an error in an upper bound\par\nobreak
Another AMO Bench problem gives $2025$ real numbers $x_1,\ldots,x_{2025}$ in $[0,1]$ and asks for a function $f$ such that $\sum_k f(x_k)$ estimates $(\sum_k x_k)^2$ with the smallest possible worst-case absolute error. After the first correction, DeepSeek Flash's solution still claimed that its function had an error of at most 1,025,156. On rechecking, AGPV found that setting every $x_k$ to 1 gives an error of 2,049,300, exceeding the claimed bound. The corrector redid the derivation and obtained the correct answer, 512,578, which passed the final check.
\par\endgroup

\Needspace{6\baselineskip}
\par\begingroup\noindent\ignorespaces
Rechecking affected branches when the final answer stays the same\par\nobreak
An AIME 2024 problem asks for the smallest positive integer $m$ for which $17^2=289$ divides $m^4+1$, once the prime has been identified as 17. The original solution considered the four possible residues of $m$ modulo 17 separately and obtained $110,138,151,179$. AGPV found that 138 and 151 failed the divisibility condition and replaced them with 155 and 134, respectively. The minimum remained 110. In the next round, the verifier rechecked each of the four branches under the updated plan. It enumerated one full period modulo 289 in Python and confirmed all four values.
\par\endgroup

\Needspace{8\baselineskip}
\subsubsection{Case Study: Correcting an Error Caused by Incomplete Counting Coverage}

We examine the DeepSeek-V4-Pro (0813) run on \texttt{hmmt\_feb\_2025\_10}. The following is a condensed account of its saved verification, feedback, and correction trace, not a verbatim transcript.

\begin{breakablecasebox}
\small
\textbf{Problem and Initial Solution.}

\textbf{Problem.}
Compute the number of ways to arrange \(1,2,3,4,5,6,7\) around a circle such that every adjacent product is at most \(20\). Rotations and reflections count as different arrangements.

\textbf{Answer Status.}
The initial extracted answer is \(28\); the reference answer is \(56\). ProgCo and ProgCo-Py both return \(28\) on this model--problem instance.

\textbf{Initial Reasoning.}
The initial solution correctly observes that \(7\) can be adjacent only to \(1,2\), and \(6\) only to \(1,2,3\). It fixes \(7\), considers the two orientations of its neighbors, and counts only two arrangements per orientation. The error is the assertion that \(6\) cannot be adjacent to \(1\). This excludes feasible placements and leads to \(7\cdot(2+2)=28\).

\textbf{Generated Verification Plan.}
The saved plan constructs the complete set of arrangements satisfying the adjacent-product condition, computes its cardinality \(N\), and checks the structural claim that the neighbors of \(7\) are exactly \(1,2\). Verification requires both the structural claim and agreement between \(N\) and the candidate count.
\end{breakablecasebox}

\Needspace{5\baselineskip}
\begin{breakablecasebox}
\small
\textbf{First Verification Cycle: Checking the Count.}

\textbf{Tool Task.}
The verifier requests the complete count and the structural check together. Ordered permutations \(p=(p_1,\ldots,p_7)\) are retained exactly when
\[
p_i p_{i+1}\leq20\quad(i=1,\ldots,6),
\qquad p_7p_1\leq20.
\]
Counting ordered permutations respects the problem's rule that rotations and reflections are distinct. The computation can be summarized as:
\begin{quote}
\footnotesize\ttfamily
V = permutations satisfying every adjacent-product bound\\
N = len(V)\\
Claim7 = every sequence in V has neighbors(7) = \{1, 2\}\\
return (N, Claim7)
\end{quote}

\textbf{Returned Evidence and Diagnosis.}
The saved Python result is \(\{N:56,\ \mathrm{Claim7}:\mathrm{True}\}\). Thus, the structural constraint is supported but the candidate count \(28\) fails. Verification returns False. The feedback retains the correct adjacency observation and directs correction to recompute the count after fixing \(7\), rather than treating the incomplete set as exhaustive.
\end{breakablecasebox}

\begin{breakablecasebox}
\small
\textbf{Feedback-Guided Correction and Reverification.}

The correction stage enumerates the arrangements with \(7\) fixed, obtains eight possibilities, and identifies the omitted placements where \(6\) is adjacent to \(1\). Its revised argument is as follows.

For the orientation
\[
7-1-a-b-c-d-2-7,
\]
the four remaining values are \(3,4,5,6\). The value \(6\) can occupy either endpoint \(a\) or \(d\). If \(a=6\), then \(b=3\), with \(4,5\) in either order. If \(d=6\), then \(c=3\), again with \(4,5\) in either order. The four valid arrangements are
\[
\begin{aligned}
&(7,1,6,3,4,5,2),\quad (7,1,6,3,5,4,2),\\
&(7,1,4,5,3,6,2),\quad (7,1,5,4,3,6,2).
\end{aligned}
\]
The other orientation of \(1,2\) around \(7\) contributes four further arrangements. Restoring the seven rotational positions gives
\[
7\cdot(4+4)=56.
\]
No division by two is made because reflections are distinct. The first cycle therefore revises the answer from \(28\) to \(56\); the next cycle verifies the revised answer and terminates successfully.

\textbf{What This Case Demonstrates.}
The trace illustrates using a complete-target computation to reject an incomplete count, retaining valid structural evidence, and directing the subsequent correction to the missing cases. It also shows that the verifier's diagnosis and the correction agent's more detailed explanation are distinct parts of the process.
\end{breakablecasebox}

\Needspace{12\baselineskip}
\par\begingroup\noindent\ignorespaces
\subsection[Component ablations]{Component ablations}\label{app:zh:ablation}
Table~\ref{tab:agpv_added_ablation} reports component ablations on 170 problems using DeepSeek Flash\textsuperscript{*}.\footnote{By the time we ran these tests, DeepSeek had updated its API and the Flash version used for the main experiments could no longer be selected. DeepSeek Flash\textsuperscript{*} denotes the updated version. DeepSeek Flash denotes the version before the update.} Full AGPV and its six variants started from the same initial answers. Each variant changed one design choice.

\noindent\begin{minipage}{\linewidth}
\captionof{table}{Component ablations on DeepSeek Flash\textsuperscript{*}}\label{tab:agpv_added_ablation}
{\small\setlength{\tabcolsep}{4pt}\renewcommand{\arraystretch}{1.08}
\begin{tabularx}{\linewidth}{@{}Xrrr@{}}\toprule
Method & Accuracy & \shortstack{Model calls\\per problem} & \shortstack{Total tokens\\(k/problem)} \\ \midrule
Full AGPV & 77.1\% & 19.44 & 154.37 \\
A. Continue past the first error & 74.1\% & 17.96 & 226.95 \\
B. Remove the explicit prompt to check for omissions & 72.4\% & 16.34 & 198.14 \\
C. Keep the first-round plan fixed & 76.5\% & 16.03 & 128.38 \\
D. Disable mathematical tools & 67.1\% & 6.71 & 54.57 \\
E. Verify directly without first listing key steps & 71.8\% & 18.33 & 244.66 \\
F. Omit the feedback summary & 75.9\% & 16.28 & 193.23 \\
\bottomrule\end{tabularx}\par}
\end{minipage}\par
\par\smallskip
Early stopping at the first error, explicit checks for omissions, and plan updates (A--C). AGPV moves directly to correction at the first error and need not read the remaining derivation. Without early stopping, it used an additional 72.58k tokens per problem, including 69.24k additional input tokens. On a problem asking for all integer solutions, full AGPV returned the correct set, $\{1,2,3\}$. Without the prompt to check for omissions, the answer also included 4 and 5. AGPV updates the plan so that the next round checks the revised reasoning. Accuracy was 76.5\% when the first-round plan was held fixed.
\par\smallskip
Mathematical tools, verification plan, and feedback summary (D--F). Full AGPV can use Python to check intermediate results and find and correct calculation errors in the original solution. Variant D, with the mathematical tools disabled, achieved 67.1\% accuracy. Full AGPV first lists the key steps and then checks each one. On some problems, variant E skipped the plan and repeatedly read long derivations during correction, increasing mean token usage per problem. Full AGPV summarizes the error location and verification results for the corrector. Variant F reads the original records directly and can still use them to revise the solution. In the table, F achieved accuracy close to that of full AGPV but used more tokens.
\par\endgroup

\end{document}